\documentclass[10pt,journal,compsoc]{IEEEtran}

\ifCLASSOPTIONcompsoc
  \usepackage[nocompress]{cite}
\else
  \usepackage{cite}
\fi

\ifCLASSINFOpdf
\else
\fi

\usepackage{times}
\usepackage{epsfig}
\usepackage{graphicx}
\usepackage{amsmath}
\usepackage{amssymb}
\usepackage{acronym}
\usepackage[]{algorithm2e}
\usepackage{footmisc}
\usepackage{chapterbib}

\newcommand{\norm}[1]{\left\lVert#1\right\rVert}
\DeclareMathOperator*{\argmin}{arg\,min}

\acrodef{BIER}{Boosting Independent Embeddings Robustly}
\acrodef{CNN}{Convolutional Neural Network}
\acrodef{SVD}{Singular Value Decomposition}
\acrodef{SGD}{Stochastic Gradient Descent}
\acrodef{LMNN}{Large Margin Nearest Neighbor}
\acrodef{RBM}{Restricted Bolzman Machine}
\acrodef{ReLU}{Rectified Linear Unit}
\acrodef{GAN}{Generative Adversarial Network}
\acrodef{DANN}{Domain Adversarial Neural Network}
\acrodef{NCL}{Negative Correlation Learning}
\acrodef{NCA}{Neighborhood Component Analysis}

\makeatletter
\DeclareRobustCommand\onedot{\futurelet\@let@token\@onedot}
\def\@onedot{\ifx\@let@token.\else.\null\fi\xspace}

\def\eg{\emph{e.g}\onedot} 
\def\ie{\emph{i.e}\onedot} 
\def\cf{\emph{c.f}\onedot} 
\def\etc{\emph{etc}\onedot} 
\def\wrt{w.r.t\onedot} 
\def\etal{\emph{et al}\onedot}
\usepackage{color}
\definecolor{green}{rgb}{0,0.6,0}
\definecolor{red}{rgb}{1,0,0}
\definecolor{white}{rgb}{1,1,1}
\makeatother

\newcommand{\bordergreen}[1] {
    \fboxrule=2pt
    \fboxsep=0pt
    \fcolorbox{green}{white}{#1}
}

\newcommand{\borderred}[1] {
    \fboxrule=2pt
    \fboxsep=0pt
    \fcolorbox{red}{white}{#1}
}

\usepackage{amsmath}

\begin{document}

\title{Deep Metric Learning with BIER: \\ Boosting Independent Embeddings Robustly}

\author{Michael~Opitz,
        Georg~Waltner,
        Horst~Possegger,
        and~Horst~Bischof
\IEEEcompsocitemizethanks{
\IEEEcompsocthanksitem M. Opitz, G. Waltner, H. Possegger and H. Bischof are with the Institute of Computer Graphics and Vision, Graz University of Technology, Austria, 8010 Graz.
\protect\\
E-mail: \{michael.opitz, waltner, possegger, bischof\}@icg.tugraz.at
}
\thanks{Manuscript received January 15, 2018.}}

\markboth{TPAMI Submission}%
{Opitz \MakeLowercase{\textit{et al.}}: BIER: Boosting Independent Embeddings Robustly}

\IEEEtitleabstractindextext{%
\begin{abstract}
Learning similarity functions between image pairs with deep neural networks yields highly correlated activations of embeddings.
In this work, we show how to improve the robustness of such embeddings by exploiting the independence within ensembles.
To this end, we divide the last embedding layer of a deep network into an embedding ensemble and
formulate training this
ensemble as an online gradient boosting problem. Each learner receives a
reweighted training sample from the previous learners. Further, we propose two loss functions 
which increase the diversity in our ensemble. These loss functions can be applied either for weight 
initialization or during training. Together, our contributions leverage large
embedding sizes more effectively by significantly reducing correlation of
the embedding and consequently increase retrieval accuracy of the embedding. Our method works with any differentiable 
loss function and does not introduce any additional parameters during test time. 
We evaluate our metric learning method
on image retrieval tasks and show that it improves over
state-of-the-art methods on the CUB-200-2011, Cars-196, Stanford Online
Products, In-Shop Clothes Retrieval and VehicleID datasets.
\end{abstract}

}

\maketitle

\IEEEdisplaynontitleabstractindextext

\IEEEpeerreviewmaketitle

\IEEEraisesectionheading{\section{Introduction}\label{sec:introduction}}

\IEEEPARstart{D}{eep} \ac{CNN} based metric learning methods map images to a high dimensional feature space. In this space semantically
similar images should be close to each other, whereas semantically dissimilar images should be far apart from each other.
To learn such metrics, several approaches based
on image pairs (\eg~\cite{chopra2005contrastive, hadsell2006dimensionality}),
triplets (\eg~\cite{schroff2015facenet,weinberger2009distance}) or
quadruples (\eg~\cite{law2013quadruplet,zheng2013reid}) have been proposed in the
past.
Metric learning has a variety of applications, such as image or object retrieval
(\eg~\cite{oh2016deep, ustinova2016histogram, wohlhart2015learning}), single-shot object classification (\eg~\cite{oh2016deep,
ustinova2016histogram, waltner2016}), keypoint descriptor learning (\eg~\cite{kumar2016learning, simo2015discriminative}), 
face verification
(\eg~\cite{parkhi2015deep, schroff2015facenet}), person re-identification
(\eg~\cite{shi2016, ustinova2016histogram}),  object
tracking (\eg~\cite{Tao_2016_CVPR}),~\emph{etc}. 

In this work, we focus on learning simple similarity functions based on the dot product, since they can be computed rapidly 
and thus, facilitate
approximate search methods (\eg~\cite{muja2014scalable}) for large-scale image retrieval. 
Typically, however, the accuracy of these methods saturates or declines due to over-fitting, especially when large embeddings are used~\cite{oh2016deep}.

\begin{figure}[t]
    \begin{center}
        \includegraphics[width=0.5\textwidth]{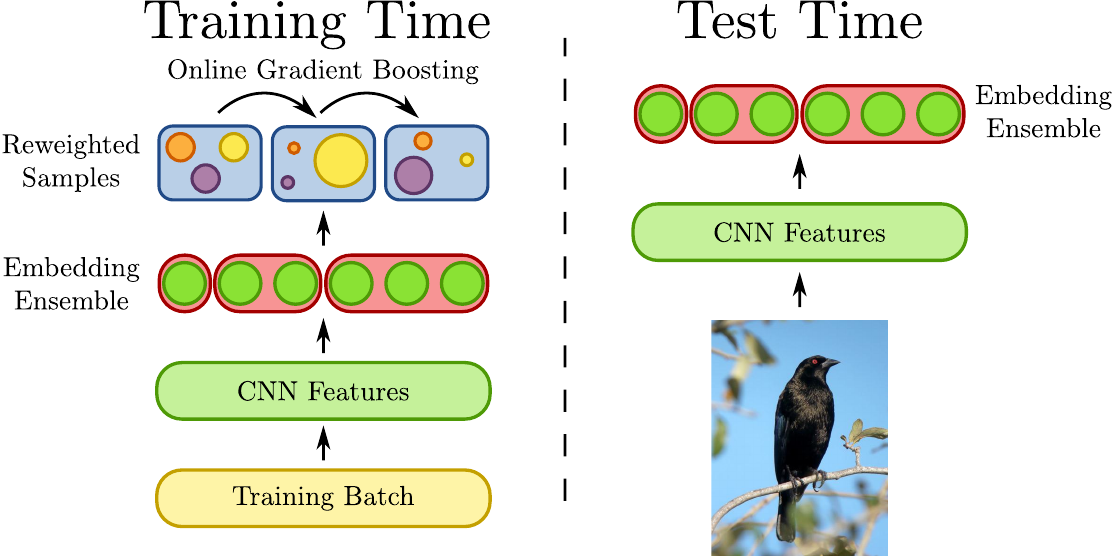}
    \end{center}
    \caption{BIER divides a large embedding into 
    an ensemble of several smaller embeddings. During training we reweight the training set for successive learners in the ensemble with the negative gradient of the loss function.
    During test time we concatenate the individual embeddings of all learners into a single embedding vector.}
    \label{fig:motivation}
\end{figure}

To address this issue, 
we present a learning approach, called \ac{BIER}, which 
leverages large embedding sizes more effectively. The main idea is to divide the last embedding
layer of a \ac{CNN} into multiple non-overlapping groups (see Fig.~\ref{fig:motivation}). Each group
is a separate metric learning network on top of a shared feature representation. 
The accuracy of an ensemble depends on the accuracy of individual learners
as well as the correlation between them~\cite{breiman2001random}. Ideally, individual learners are highly
accurate and have low correlation with each other, so that they complement each other during 
test time.

Na\"ively optimizing a global loss function for the whole ensemble shows no benefits since all learners have access to the same feature representation and the same training samples. All
groups will end up learning highly correlated embeddings, which results in no performance improvements at all, which is especially true for metric learning.
To overcome this problem, we formulate the ensemble training as an online
gradient boosting problem. In online gradient boosting, each learner reweights a training sample for
successive learners according to the gradient of the loss function. Consequently,
successive learners will focus on different samples than the previous learners,
resulting in a more diverse feature representation (Section~\ref{sec:gradient-boosting}). 
To encourage the individual embeddings to have low correlation with each other already at the beginning of the training, 
we propose a novel initialization method for our embedding matrix (Section~\ref{sec:diversity-measurements} and Section~\ref{sec:optimizing-diversity}). 
The matrix is initialized from a solution of an optimization problem which implicitly minimizes
the correlation between groups. 

In comparison to our earlier version of this work~\cite{opitz2017bier}, we
extend \ac{BIER} by integrating our weight initialization method as auxiliary loss function directly into
the training objective (Section~\ref{sec:method-auxiliary-loss}). As we show in our evaluation
(Section~\ref{sec:eval-auxiliary-loss-function}), this allows training \ac{BIER} at higher learning rates
which significantly reduces training time. By jointly training our network with this loss function, we can further reduce the correlation between 
learners and improve
the accuracy of our method (Section~\ref{sec:eval-auxiliary-loss-function}).

Additionally, we improve our the performance by introducing a novel Adversarial Loss, which learns adversarial regressors
between pairs of embeddings (Section~\ref{sec:adversarial-loss}). These 
regressors learn a non-linear transformation between embeddings. Their objective is to maximize similarity 
between embeddings.
Between
our embeddings and the regressors, we insert a gradient reversal
layer~\cite{ganin2016domain}. This layer changes the sign of the gradients during backpropagation and behaves like 
the identity function during forward propagation.
As a consequence, our embeddings are trained to maximize this loss function \wrt our adversarial regressors and hence our ensemble becomes even more diverse.

We demonstrate the effectiveness of our metric on several image retrieval
datasets~\cite{krause20133d, liu2016deep, liu2016deepfashion, oh2016deep, WahCUB_200_2011}.
In our evaluation we show that \ac{BIER} 
significantly reduces the correlation of large embeddings (Section~\ref{sec:eval:strength-and-correlation}) 
and works with several loss functions (Section~\ref{sec:eval-loss-functions}) 
while increasing retrieval accuracy by a large margin.
\ac{BIER} does not introduce any additional parameters into a \ac{CNN} and has only negligible additional cost during training time and runtime. 
We show that \ac{BIER} achieves
state-of-the-art performance on the
CUB-200-2011~\cite{WahCUB_200_2011}, Cars-196~\cite{krause20133d}, Stanford
Online Products~\cite{oh2016deep}, In-Shop Clothes Retrieval~\cite{liu2016deepfashion} and VehicleID~\cite{liu2016deep} datasets (Section~\ref{sec:eval-sota}).
Further, by employing our novel Adversarial Loss during training time as auxiliary loss, we can significantly outperform the state-of-the-art on these datasets.

\section{Related Work}
\label{sec:related-work}

Our work is related to metric learning (Section~\ref{sec:related-work-metric-learning}) and boosting in
combination with \acp{CNN} (Section~\ref{sec:related-work-cnn-boosting}). Additionally, since we propose a novel initialization
method, we discuss related data dependent initialization methods for
\acp{CNN} (Section~\ref{sec:related-work-init}). 
Next, we discuss techniques to increase the diversity of ensembles related to our auxiliary function (Section~\ref{sec:related-work-diversity}). 
Finally, we summarize adversarial loss functions for \acp{CNN} (Section~\ref{sec:related-work-adversarial}), 
as we use an adversarial loss to encourage diversity of our learners.

\subsection{Metric Learning}
\label{sec:related-work-metric-learning}

The main objective of metric learning in Computer Vision is to learn a distance
function $d(\cdot, \cdot): \mathbb{R}^k \times \mathbb{R}^k \mapsto \mathbb{R}^+$
mapping two $k$-dimensional input vectors, which are typically an input image
or a feature representation of an image, to a distance between images. 
Typically, these
distance functions have the form ${d(\boldsymbol{x}, \boldsymbol{y})^2 =
(\boldsymbol{x} - \boldsymbol{y})^\top \boldsymbol{M} (\boldsymbol{x} -
\boldsymbol{y})}$, where $\boldsymbol{M}$ is a positive semidefinite matrix. $\boldsymbol{M}$ can be
factorized as  ${(\boldsymbol{x} - \boldsymbol{y})^\top
\boldsymbol{L}\boldsymbol{L}^{\top} (\boldsymbol{x} - \boldsymbol{y}) =
\norm{\boldsymbol{x^{\top}L} - \boldsymbol{y^{\top}L}}^2}$, where $\boldsymbol{L} \in
\mathbb{R}^{k \times d}$ projects an image, or a feature representation of an
image into a $d$-dimensional vector space. In this vector space, semantically similar
images should be close to each other, whereas semantically dissimilar images should be far apart from each other.

For a complete review of metric learning approaches we refer the interested reader to~\cite{bellet2013survey}. 
In this work we focus our discussion on boosting based metric learning approaches and deep \ac{CNN} based approaches.

\subsubsection{Boosting Based Metric Learning}
In boosting based approaches, weak learners are typically rank one matrices. The
ensemble then combines several of these matrices to form a positive semidefinite
matrix $\boldsymbol{M}$, \eg~\cite{bi2011adaboost, liu2012robust, negrelbmvc16,
shen2012positive}. 
Kedem~\etal~\cite{kedem2012non} propose gradient boosted trees for metric learning. They learn the non-linear mapping 
$f(\cdot)$ with an ensemble of regression trees, by minimizing a \ac{LMNN} loss function~\cite{weinberger2009distance} with the gradient boosting framework.
Further, they initialize their first learner as the solution of the linear \ac{LMNN} optimization problem.
In contrast to these offline boosting based works, our method is an online boosting method, which directly integrates into deep \ac{CNN} training.
Our weak learners are fully connected layers on top of a shared \ac{CNN} feature representation and, compared to these methods, typically have a higher rank. Further, we use auxiliary loss
functions to explicitly encourage diversity in our metric ensemble.

\subsubsection{CNN Based Metric Learning}
\ac{CNN} based methods learn a non-linear transformation of an input image of 
the form $\phi(\cdot): \mathbb{R}^k \mapsto \mathbb{R}^h$. This \ac{CNN} based 
feature extractor, \ie $\phi(\cdot)$, can be pre-trained on 
other tasks, such as large scale image classification, \eg~\cite{ILSVRC15}, and is then 
fine-tuned on metric learning datasets.
To map the feature
representation into the $d$-dimensional vector space, an additional linear embedding layer
is typically added at the end of a \ac{CNN} feature extractor 
as ${f(\boldsymbol{x}) = \phi(\boldsymbol{x})^{\top}\boldsymbol{W}}$, ${\boldsymbol{W} \in \mathbb{R}^{h \times d}}$.
Hence, metric learning \acp{CNN} learn the distance function
 ${d(\boldsymbol{x}, \boldsymbol{y})^2 = (\phi(\boldsymbol{x}) - \phi(\boldsymbol{y}))^{\top}\boldsymbol{W}\boldsymbol{W}^{\top}(\phi(\boldsymbol{x}) - \phi(\boldsymbol{y}))}$,
which is equivalent to $(\phi(\boldsymbol{x}) - \phi(\boldsymbol{y}))^{\top}\boldsymbol{M}(\phi(\boldsymbol{x}) - \phi(\boldsymbol{y}))$.
To jointly learn all parameters of the \ac{CNN} and the embedding, special loss functions operating on image pairs, 
triplets or quadruples are used.
One of the most widely used pairwise loss functions for metric learning is the contrastive 
loss function, \eg~\cite{chopra2005contrastive, hadsell2006dimensionality, oh2016deep}. This loss function 
minimizes the squared Euclidean distance between positive feature vectors while encouraging
a margin between positive and negative pairs. To train networks with this loss function, a Siamese architecture, \ie two copies
of a network with shared weights, is commonly used, \eg~\cite{chopra2005contrastive, hadsell2006dimensionality}.

Other approaches adopt the \ac{LMNN} formulation~\cite{weinberger2009distance}
and sample triplets consisting of a positive image pair and a negative image
pair, \eg~\cite{oh2016deep, parkhi2015deep, schroff2015facenet, wohlhart2015learning}.  The loss function encourages a margin
between distances of positive and negative pairs. Hence, positive image pairs
are mapped closer to each other in the feature space compared to negative 
image pairs.

Recently, several new loss functions for metric learning have been proposed. 
Song~\etal~\cite{oh2016deep} propose to lift a mini-batch to a matrix of pairwise distances between 
samples. They use a structural loss function on this distance matrix to train the neural network.
Ustinova~\etal~\cite{ustinova2016histogram} propose a novel histogram loss. They also lift a
mini-batch to a distance matrix and compute a histogram of positive and negative distances. Their loss operates on 
this histogram and minimizes the overlap between the distribution of positive and negative distances.
Huang~\etal~\cite{huang2016local} introduce a position dependent deep metric unit which 
is capable of learning a similarity metric adaptive to the local feature space.
Sohn~\cite{sohn2016improved} generalizes the triplet loss to n-tuples and propose a more efficient 
batch construction scheme.
Song~\etal~\cite{song2017cvpr} propose a structured clustering loss to train embedding networks.
Wang~\etal~\cite{wang2017iccv} propose a novel angular loss, which improves the traditional triplet loss by imposing geometric constraints for triplets.
Movshovitz-Attias~\etal~\cite{movshovitz-attias2017iccv} propose a proxy-loss where they introduce a set of proxies which approximate the dataset. 
Their Proxy-\ac{NCA} loss function optimizes distances to these proxies. 
Rippel~\etal~\cite{rippel2015metric} propose a ``magnet'' loss function which models multimodal data distributions and
minimizes the overlap between distributions of different classes.

Our work is complementary to these approaches. 
We show in our evaluation that combining existing loss functions with our 
method yields significant improvements (Section~\ref{sec:eval-loss-functions}).

Another line of work aims at improving the sample mining strategy used for
embedding learning. Schroff~\etal~\cite{schroff2015facenet} propose a
semi-hard mining strategy for the triplet loss. Within a mini-batch, they only
use samples for training where the negative image pair has a larger distance
than the positive pair. This avoids getting stuck in a local minima early in
training \cite{schroff2015facenet}.  Harwood~\etal~\cite{harwood2017iccv} use
offline sampling of training samples. To avoid the large computational cost,
they use approximate nearest neighbor search methods to accelerate distance
computation. Wu~\etal~\cite{wu2017iccv} propose a distance weighted sampling
method in combination with a margin based loss function to improve metric
learning.

Although the main objective of our method is to reduce correlation in a large
embedding, we apply a form of hard negative mining. We reweight samples for
successive learners according to the gradient of the loss function. More
difficult samples are typically assigned a higher gradient than easier samples.
Hence, successive learners focus on harder examples than previous learners. 
However, we do not use any sample mining strategy for our first learner and hypothesize that our 
method can benefit from the above approaches, \eg by selecting 
better samples from the training-set or mini-batch.

Most closely related to our method is the concurrent work of Yuan~\etal~\cite{yuan2016hard}. 
They propose a hard-aware deeply cascaded embedding.
This method leverages the benefits of deeply supervised
networks~\cite{lee2015deeply,Szegedy2014} by employing a contrastive loss
function and train lower layers of the network to handle easier examples, and
higher layers in a network to handle harder examples. In contrast to this
multi-layer approach, we focus on reducing the correlation on just a single
layer.  
Further, our method
allows continuous weights for samples depending on the loss function.
Finally, we show that 
employing auxiliary loss functions during initialization or training decreases correlation of learners
and consequently improves the accuracy of the ensemble.

\subsection{Boosting for CNNs}
\label{sec:related-work-cnn-boosting}

Boosting is a greedy ensemble learning method, which iteratively trains an
ensemble from several weak learners~\cite{freund1997boost}. The original boosting algorithm, AdaBoost~\cite{freund1997boost}, minimizes an exponential loss function.  
Friedman~\cite{friedman2001} extends the boosting framework to allow minimizing arbitrary differentiable loss functions. 
They show that one interpretation of boosting is that it
performs gradient descent in function space and propose a novel method leveraging this insight called
gradient boosting. 
Successive learners in gradient boosting
are trained to have high correlation with the negative gradient of
the loss function. There are several algorithms which extend gradient boosting
for the online learning setting, \eg~\cite{beygelzimer2015online,
beygelzimer2015optimal,  chen2012online, leistner2009robustness}. In contrast
to offline boosting, which has access to the full dataset, online boosting
relies on online weak learners and updates the boosting model and their weak
learners one sample at a time. 

In the context of \acp{CNN} these methods are rarely
used. Several works, \eg~\cite{karianakis2015boosting, yang2015convolutional}
use \ac{CNN} features in an offline boosting framework. These approaches,
however, do not train the network and the weak learners end-to-end, \ie the
\ac{CNN} is typically only used as a fixed feature extractor. In contrast to these approaches, we train our system end-to-end. 
We directly incorporate an online boosting algorithm into training a \ac{CNN}.

Similarly, Walach~\etal~\cite{walach2016learning} leverage gradient boosting to train several \acp{CNN} within an offline gradient boosting framework for person counting. 
The ensemble is then fine-tuned with a global loss function. In contrast to their work, which trains several copies of full \ac{CNN} models, 
our method trains a single \ac{CNN} with an online boosting method. Similar to dropout~\cite{srivastava14dropout}, all our learners share a common feature representation. Hence, 
our method does not introduce any additional parameters. 

Very recently, Han~\etal~\cite{han2016actionunit} propose to use boosting to select
discriminative neurons for facial action unit classification. They employ
decision stumps on top of single neurons as weak learners, and learn weighting
factors for each of these neurons by offline AdaBoost~\cite{freund1997boost} applied to each
mini-batch separately. Weights are then exponentially averaged over several
mini-batches. They combine the weak learner loss functions with a global loss
function over all learners to train their network. In contrast to this work, we
use weak learners consisting of several neurons (\ie linear classifiers).
Further, our method is more tightly integrated in an online boosting
framework.  We reweight the training set according to the negative gradient of
the loss function for successive weak learners.  This encourages them to focus
on different parts of the training set. Finally, our method does not rely on
optimizing an explicit discriminative global loss function. 

\subsection{Initialization Methods}
\label{sec:related-work-init}

Most initialization methods for \acp{CNN} initialize weights randomly, either
with carefully chosen variance parameters, \eg~\cite{krizhevsky2012imagenet}, or depending
on the fan-in and fan-out of a weight matrix, \eg~\cite{Glorot10, He2015}, with the goal of having an 
initialization which provides a large gradient during learning.
Rather than focusing on determining the variance of the weight matrix, Saxe~\etal~\cite{SaxeMG13}
propose to initialize the weight matrix as orthogonal matrix.

Recently, several approaches which initialize weights depending on
the input data were proposed, \eg~\cite{krahenbuhl2015data, mishkin2016all}. These methods
typically scale a random weight matrix such that the activations on the training set have 
unit variance.

Another line of work, \eg~\cite{bengio2006greedy, hinton2006fast}, 
greedily initializes a network layer-by-layer, by applying unsupervised
feature learning, such as Autoencoders or \acp{RBM}.
These methods seek for a weight matrix which minimizes the reconstruction error or a matrix which
learns a generative model of the data.

Our initialization method is also a form of unsupervised pre-training of a single layer, as 
we use unsupervised loss functions for initializing the weights of our embedding layer.
However, as opposed to minimizing a reconstruction loss or learning a generative model of the data,
we initialize the weight matrix from a solution of an
optimization problem which implicitly minimizes correlation between groups of features.
With this initialization our weak learners already have low correlation at the beginning
of the training process.

\subsection{Diversity in Ensembles}
\label{sec:related-work-diversity}

Previous approaches which exploit diversity in ensembles are based on
\ac{NCL}~\cite{liu1999ensemble}, \eg~\cite{liu1999ensemble,chen2010multiobjective}. These methods train neural networks in an ensemble to be
negatively correlated to each other by penalizing the cross-correlation of their predictions. As a consequence,
they complement each other better during test time. These approaches are typically focused on 
training regressor ensembles, as opposed to classification or metric ensembles and do not use boosting. Further, they train several full 
regressor networks from scratch as opposed to using a single shared feature extractor \ac{CNN}.

More closely related is AdaBoost.NC~\cite{wang2010negative}, which extends \ac{NCL} to AdaBoost for classification. AdaBoost.NC
defines an ambiguity penalty term based on the deviation of the predictions of the weak learners to the ensemble
prediction. Intuitively, if many learners deviate from the ensemble prediction for a sample, the ambiguity is high.
This ambiguity measure is used to update the weights for the samples for successive learners in the ensemble.
In contrast to this work, we encourage diversity in our ensemble by directly using a differentiable loss function for our learners.

Finally, in an earlier work we applied auxiliary loss functions for
a deep \ac{CNN} based classification ensemble with a shared feature
representation~\cite{opitz2016accv}. Similar to this work, for computational efficiency, we share all low level \ac{CNN} features 
and divide the network at the end into several non-overlapping groups. In contrast to our earlier work, we use online boosting to build
our metric ensemble and different loss functions which are compatible with metric
learning to encourage diversity.

\subsection{Adversarial Loss Functions}
\label{sec:related-work-adversarial}

Adversarial networks, such as \acp{GAN}~\cite{goodfellow2014generative}, have several applications such as image
generation (\eg~\cite{goodfellow2014generative, radford2016dcgan, mao2017least}), style transfer (\eg~\cite{CycleGAN2017}), 
domain adaptation (\eg~\cite{tzeng2017adversarial}), \etc.
These approaches typically consist of two neural networks, a discriminator and a generator. 
During training, discriminator and generator are playing a two-player minimax game.
The discriminator minimizes a loss function to distinguish real-world images from 
fake images, which are generated by the generator.
On the other hand, the generator tries to confuse the discriminator by
generating plausible fake images. To achieve this, it maximizes the loss
function the discriminator tries to minimize. The problem has a unique solution
where the generator recovers the training data distribution and the
discriminator assigns an equal probability of $\frac{1}{2}$ to real-world and
fake samples~\cite{goodfellow2014generative}.
During training, \acp{GAN} use alternating \ac{SGD} to optimize the two networks. In the first step the parameters of the generator are updated, keeping the parameters of the discriminator fixed. Then, in the second step the discriminator is updated, while keeping the generator fixed
\eg~\cite{goodfellow2014generative, CycleGAN2017, radford2016dcgan, mao2017least}.

Most closely related to our work are methods which apply \acp{GAN} and adversarial loss
functions for domain adaptation.
Tzeng~\etal~\cite{tzeng2015simultaneous} propose an adversarial loss at
feature level for domain adaptation. They train a linear classifier on top of a
hidden feature representation to categorize the domain of a sample. The feature
generator (\ie the hidden representation of the neural network) is trained to maximize the
loss function of this classifier. Consequently, the hidden
representation of samples from different domains will be aligned and hence undistinguishable for the linear classifier.

Similar to the \ac{GAN} setup, Ganin~\etal~\cite{ganin2016domain} propose \acp{DANN}. This method uses a gradient reversal layer for domain adaptation. 
They insert a discriminator on top of a neural network feature generator. The discriminator minimizes a loss function 
to distinguish samples of two different domains. Between the discriminator and feature extractor they insert a gradient reversal
layer which flips the sign of the gradients during backpropagation. As a consequence, the feature extractor maximizes the loss
function of the discriminator, making the hidden layer representation of different domains undistinguishable for the 
discriminator. Compared to \ac{GAN} based approaches, \acp{DANN} do not need alternating updates
of the generator and discriminator. At each step, the method updates the parameters of both, the generator and the discriminator.

As opposed to aligning two domains with each other, our method makes embeddings more diverse. To this end, we 
adopt the gradient reversal layer of \acp{DANN} to make different
learners more diverse from each other. We train a regressor, as opposed to a discriminator,
which projects features from one learner to the other with a non-linear neural
network. We optimize the regressor to maximize the similarity between embeddings. By inserting the gradient reversal
layer between the regressor and our embeddings, we force our embeddings to be more diverse to each other. 
To the best of our knowledge, domain adaptation approaches have not been applied to increase diversity among classifiers.

\section{Boosting a Metric Network}

\begin{figure}[t]
    \begin{center}
        \includegraphics[width=0.24\textwidth]{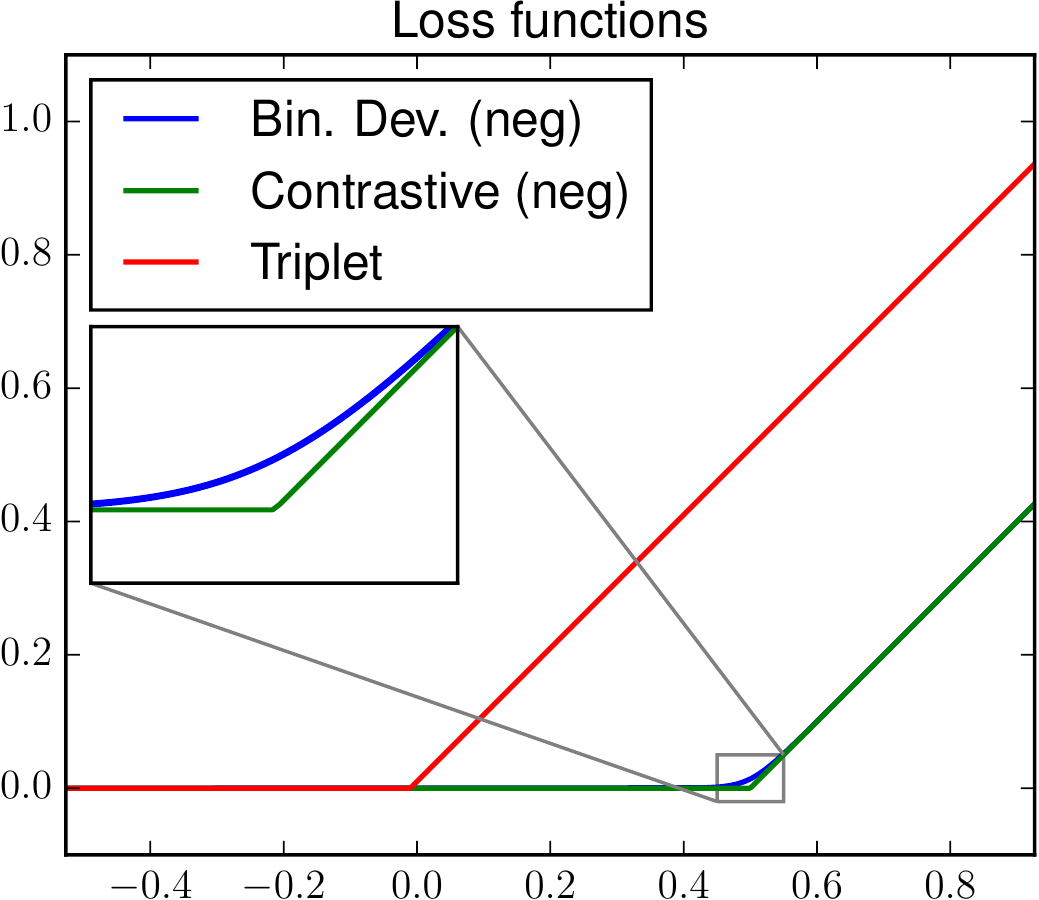}
        \includegraphics[width=0.24\textwidth]{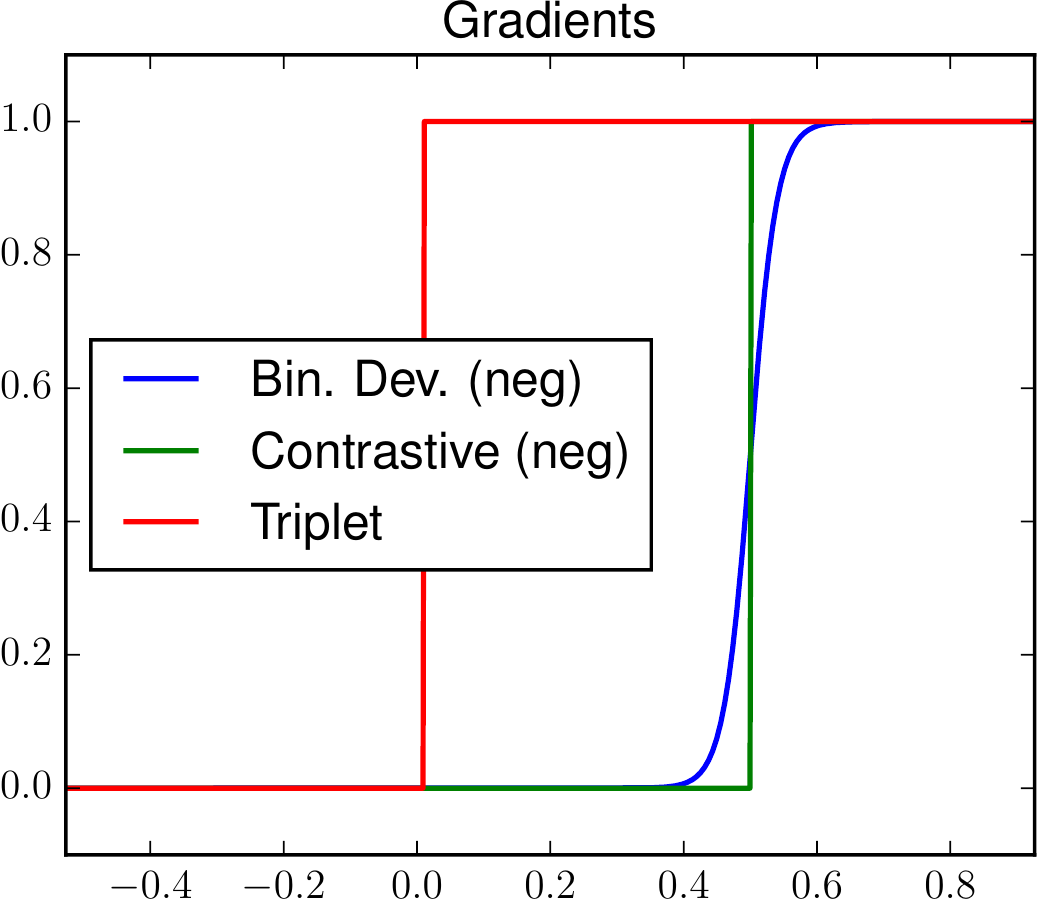}
    \end{center}
    \caption{Illustration of triplet loss, contrastive loss (for negative samples) and binomial deviance loss (for negative samples) and their gradients.
             Triplet and contrastive loss have a non-continuous gradient, whereas binomial deviance has a continuous gradient.}
    \label{fig:gradients}
\end{figure}

\begin{table}[t]
    \caption{Definition of loss functions used in our work.}
    \label{tbl:loss-function-overview}
    \renewcommand{\arraystretch}{1.3}
    \centering
    \begin{tabular}{ll}
        \hline
        Binomial Deviance & $\log(1 + e^{-(2 y - 1) \beta_1 (s - \beta_2) C_y })$ \\
        Contrastive   & $(1 - y) \max(0, s - m) + y (s - 1)^2$ \\
        Triplet       & $\max(0, s^- - s^+ + m)$ \\
        \hline
    \end{tabular}
\end{table}

Our method builds upon metric \acp{CNN}, \eg~\cite{huang2016local, oh2016deep, sohn2016improved, ustinova2016histogram}. The main objective of these networks is 
to learn a high-dimensional non-linear embedding $f(\boldsymbol{x})$, which maps an image $\boldsymbol{x}$ to a feature space $\mathbb{R}^d$. 
In this space, similar image pairs should be close
to each other and dissimilar image pairs should be far apart from each other. To achieve this, instead of 
relying on a softmax output layer, these methods
use a final linear layer consisting of an embedding matrix $\boldsymbol{W} \in \mathbb{R}^{h \times d}$, which maps 
samples from the last hidden layer of size $h$ into the feature space $\mathbb{R}^d$. 
To learn this embedding matrix $\boldsymbol{W}$ and the parameters of the underlying network, 
these networks are typically trained on pairs or triplets of images and use loss functions
to encourage separation of positive and negative pairs, \eg~\cite{oh2016deep}.

As opposed to learning a distance metric, in our work we learn a cosine similarity score $s(\cdot,
\cdot)$, which we define as dot product between two embeddings
\begin{equation}
s(f(\boldsymbol{x}^{(1)}), f(\boldsymbol{x}^{(2)})) = \frac{f(\boldsymbol{x}^{(1)})^{\top}f(\boldsymbol{x}^{(2)})}{\norm{f(\boldsymbol{x}^{(1)})} \cdot \norm{f(\boldsymbol{x}^{(2)})}}. 
\end{equation}
This has the
advantage that the similarity score is bounded between $\left[-1,+1\right]$.

In our framework, we do not use a Siamese architecture, \eg as~\cite{chopra2005contrastive, hadsell2006dimensionality}. Instead,
we follow recent work, \eg~\cite{oh2016deep, schroff2015facenet, ustinova2016histogram}, and 
sample a mini-batch of several images, forward propagate them through the network and sample pairs or triplets
in the last loss layer of the network. The loss is then backpropagated through all layers of the network. This has the advantage
that we do not need to keep several separate copies of the network in memory and that we can improve the computational efficiency. 

We consider three different loss functions (defined in Table~\ref{tbl:loss-function-overview} and illustrated in Fig.~\ref{fig:gradients}), which are commonly used to train metric networks,
\eg~\cite{kumar2016learning, parkhi2015deep, schroff2015facenet, shi2016}. 
To avoid cluttering the notations in Table~\ref{tbl:loss-function-overview}, let $s = s(f(\boldsymbol{x}^{(1)}), f(\boldsymbol{x}^{(2)}))$ 
be the similarity score between image $\boldsymbol{x}^{(1)}$ and $\boldsymbol{x}^{(2)}$. Let $y \in \{1, 0\}$ denote the label of 
the image pair (\ie $1$ for similar pairs, and $0$ for dissimilar pairs). Let $s^-$ denote the similarity score for a negative
image pair and $s^+$ denote the similarity score for a positive image pair. Further, $m$ denotes the margin for the contrastive and triplet loss, which 
is set to $0.5$ and $0.01$, respectively. $\beta_1$ and $\beta_2$ are scaling and translation parameters and are set to 
$2$ and $0.5$, similar to~\cite{ustinova2016histogram}. Finally, we follow~\cite{ustinova2016histogram} and set the cost $C_y$ to balance positive and negative pairs for the binomial deviance loss as
\begin{equation}
C_y = \left\{
    \begin{array}{c l}
        1  & \text{if}\; y = 1 \\
        25 & \text{otherwise}.
    \end{array} \right.
\end{equation}
The binomial deviance loss is similar to the contrastive loss, but has a smooth gradient (see Fig.~\ref{fig:gradients}).
In contrast, the contrastive and triplet loss have a gradient of either $0$ or $1$.
As we show in our evaluation (Section~\ref{sec:eval-loss-functions}) the binomial deviance loss benefits more from our method compared 
to the triplet and contrastive loss. We hypothesize that the main reason for that is that the gradient of the binomial deviance loss is 
smooth compared to the triplet loss or the contrastive loss.
As a consequence,
our method assigns smooth weights to training samples which conveys more information for successive learners.

\subsection{Online Gradient Boosting CNNs for Metric Learning}
\label{sec:gradient-boosting}

\begin{figure}[t]
    \begin{center}
        \includegraphics[width=0.5\textwidth]{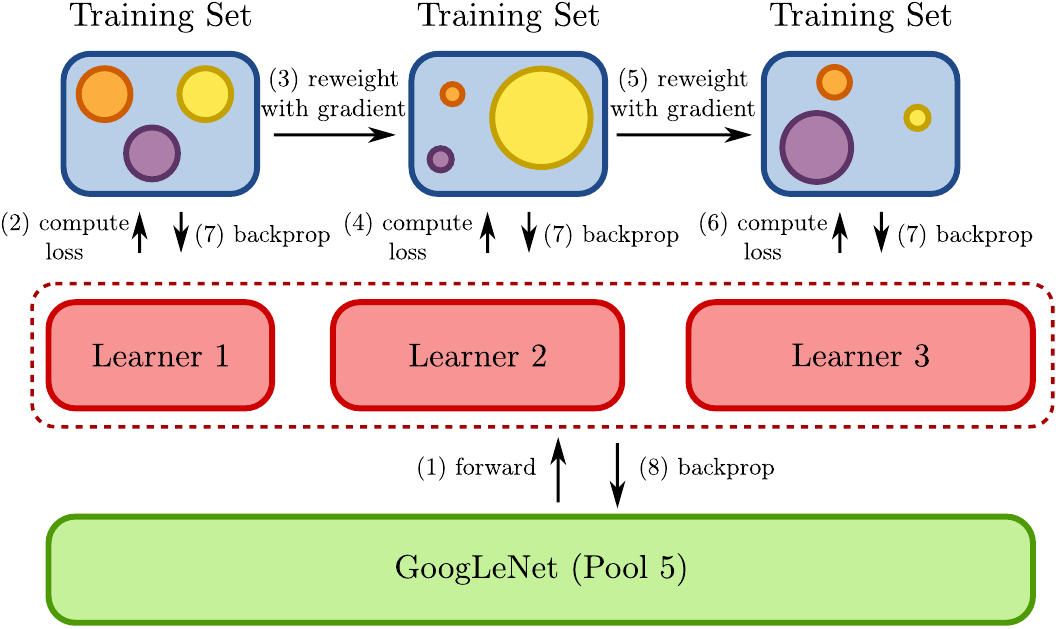}
    \end{center}
    \caption{We divide the embedding (shown as dashed layer) of a metric \ac{CNN} into several weak learners and cast training them as online gradient boosting problem.
    Each learner iteratively reweights samples according to the gradient of the loss function. 
    Training a metric \ac{CNN} this way encourages successive learners to focus on different samples than previous learners and consequently reduces correlation between learners and their feature representation.
}
    \label{fig:overview}
\end{figure}

To encourage diverse learners we borrow ideas from online gradient boosting.
Online gradient boosting iteratively minimizes a loss function using a fixed
number of $M$ weak learners, \eg~\cite{beygelzimer2015online,
beygelzimer2015optimal,  chen2012online, leistner2009robustness}. 
Learners are trained on reweighted samples according to the gradient of the
loss function. Correctly classified samples 
typically receive a lower weight while misclassified samples are assigned a higher weight for successive learners. 
Hence, successive learners focus on
different samples than previous learners, which consequently encourages higher
diversity among weak learners.

More formally, for a loss $\ell(\cdot)$, we want to find a set of weak learners
$\{f_1(\boldsymbol{x}), f_2(\boldsymbol{x}), \ldots, f_M(\boldsymbol{x})\}$ and their corresponding boosting model
\begin{equation}
F(\boldsymbol{x}^{(1)}, \boldsymbol{x}^{(2)}) = \sum_{m=1}^M \alpha_m s(f_m(\boldsymbol{x}^{(1)}), f_m(\boldsymbol{x}^{(2)})), 
\end{equation}
where $F(\boldsymbol{x}^{(1)}, \boldsymbol{x}^{(2)})$ denotes the ensemble output and $\alpha_m$ is the weighting factor of the $m$-th learner.
The $m$-th learner of the ensemble is trained on a reweighted training batch according to the negative gradient $-\ell'(\cdot)$ of 
the loss function at the ensemble prediction until stage $m-1$.

To train the weak learners $f_m(\cdot)$ in an online fashion, we
adapt an online gradient boosting learning algorithm~\cite{beygelzimer2015online} 
with fixed weights $\alpha_m$ and integrate it within a \ac{CNN}. Na\"ively
training multiple \acp{CNN} within the boosting framework is, however,
computationally too expensive.  To avoid this additional computational cost, we divide
the embedding layer of our \ac{CNN} into several non-overlapping groups, as
illustrated in Fig.~\ref{fig:overview}.  A single group represents a weak
learner. All our weak learners share the same underlying feature
representation, which is a pre-trained ImageNet \ac{CNN} in all our experiments.

Our network is trained end-to-end on mini-batches with \ac{SGD} and momentum. We illustrate the training 
procedure for loss functions operating on pairs and a single example per batch in Algorithm~\ref{algo:training}. 
Our algorithm also works with triplets, but for the sake of clarity we omit a detailed explanation here 
and refer the interested reader to the supplementary material.
The training procedure can be easily integrated
into the standard backpropagation algorithm, introducing only negligible additional cost, since most 
time during training is spent on computing convolutions.
First, in the forward pass we compute similarity scores $s_n^m$ for each input sample $n$ and each group $m$.
In the backward pass we backpropagate the reweighted losses for each group iteratively. The weight 
$w_n^m$ for the $n$-th sample and the $m$-th learner is computed from the negative gradient $-\ell'(\cdot)$ of the ensemble prediction until stage $m-1$. Hence, successive learners
focus on examples which have large gradients (\ie are misclassified) by previous learners.

\begin{algorithm}[htbp]
    Let $\eta_m = \frac{2}{m+1}$,  for $m = 1,2,\ldots,M$, \\
    $M$ = number of learners, 
    $I$ = number of iterations \\
    \For {$n = 1$ \KwTo $I$ } {
        \tcc{Forward pass} 
        Sample pair $(\boldsymbol{x}^{(1)}_n$, $\boldsymbol{x}^{(2)}_n)$ and corresponding label $y_n$ \\
        $s_n^0 := 0$ \\
        \For {m = 1 \KwTo M } {
             $s_n^m := (1 - \eta_m) s_n^{m-1} + \eta_m s(f_m(\boldsymbol{x}^{(1)}_n), f_m(\boldsymbol{x}^{(2)}_n))$
        }
        Predict $s_n$ = $s_n^M$ \\
        \vspace{2.5mm}
        \tcc{Backward pass} 
        $w_n^1 := 1$ \\
        \For {$m = 1$ \KwTo $M$ } {
            Backprop $w_n^m \ell(s(f_m(\boldsymbol{x}^{(1)}_n), f_m(\boldsymbol{x}^{(2)}_n)), y_n)$ \\
            $w_n^{m+1} := -\ell'(s_n^{m}, y_n)$
        }
    }
    
    \label{algo:training}
    \caption{Online gradient boosting algorithm for our \ac{CNN}.}
\end{algorithm}

This online gradient boosting algorithm yields a convex combination of weak learners $f_m(\cdot)$, $1 \le m \le M$. 
Successive learners in the ensemble typically have to focus on more complex training samples compared to previous learners and 
therefore, should have a larger embedding size.
We exploit this prior knowledge
and set the group size of learner $m$ to be proportional to its weight ${\alpha_m = \eta_m \cdot \prod_{n=m+1}^M (1-\eta_n)}$ in the boosting algorithm, where $\eta_m = \frac{2}{m + 1}$.
We experimentally verify this design choice in Section~\ref{sec:eval:strength-and-correlation}.

During test time our method predicts a single feature vector for an input
image $\boldsymbol{x}$.  We simply compute the embeddings from all
weak learners $f_1(\cdot), f_2(\cdot), \ldots f_M(\cdot)$, $L_2$-normalize each
of them individually and weight each of them according to 
the boosting weights $\alpha_m$. Finally, we concatenate all vectors to a
single feature vector, which is the embedding $f(\boldsymbol{x})$ of the input
image $\boldsymbol{x}$. As a consequence, distances between our vectors can 
be efficiently computed via dot products and hence, our vectors can be used by approximate search
methods, \eg~\cite{muja2014scalable}.

\subsection{Diversity Loss Functions}
\label{sec:diversity-measurements}

Rather than relying on boosting alone to increase the diversity in our
ensemble, we propose additional loss functions which make learners more
diverse from each other.
We present two different loss functions to encourage the diversity of learners. These can either be used for weight initialization or as auxiliary loss function during training (see Section~\ref{sec:optimizing-diversity}).
Our first loss function, which we denote as \emph{Activation Loss}, optimizes the embeddings such that
for a given sample, only a single embedding is active and all other embeddings are close to zero (see Section~\ref{sec:activation-loss}).
As second loss function, we propose an \emph{Adversarial Loss}. We train a regressor
on top of our embeddings which maps one embedding to a different embedding, maximizing their similarity. 
By inserting a gradient reversal layer~\cite{ganin2016domain} between the regressors and our embeddings, we
update our embeddings so that they minimize the similarity between each other with respect to these regressors 
which results in more diverse embeddings (see Section~\ref{sec:adversarial-loss}).

\subsubsection{Activation Loss}
\label{sec:activation-loss}

\begin{figure}[t]
\includegraphics[width=0.5\textwidth]{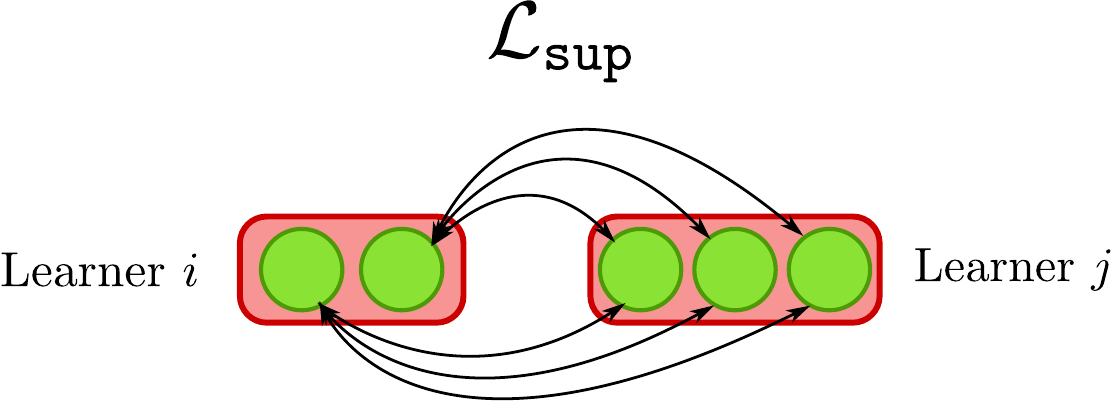}
\caption{Illustration of our Activation Loss. Neurons (green) of different embeddings (red) suppress each other. 
We apply this loss during training time between all pairs of our learners.}
\label{fig:activation-loss}
\end{figure}

Our Activation Loss directly operates on the activations of our embeddings, making our learners more diverse by suppressing
all activations except those of a single embedding (see Fig.~\ref{fig:activation-loss}). As a consequence, for a given sample, only a single embedding is active and all other embeddings 
are close to zero.
More formally, let $M$ denote the number of groups (\ie weak learners) and $G_i$ denote the index set of neurons of group 
$i$, $1 \le i \le M$. We want to increase the diversity of the embedding matrix $\boldsymbol{W} \in \mathbb{R}^{h \times d}$, 
where $d$ denotes the embedding size and $h$ the input feature dimensionality, \ie the size of the last hidden layer in a \ac{CNN}. 
Finally, let $X = \left \{ \boldsymbol{x}^{(1)}, \boldsymbol{x}^{(2)}, \ldots, \boldsymbol{x}^{(N)} \right \}$ 
denote the training set. For our initialization experiments, which we will discuss in Section~\ref{sec:initialization-method}, we use feature vectors extracted from the last 
hidden layer of a pre-trained \ac{CNN}, which we denote as $\phi(\boldsymbol{x}): \mathbb{R}^{k} \mapsto \mathbb{R}^{h}$, 
where $k$ denotes the input image dimensionality and $h$ the dimensionality of the last hidden layer of our feature extractor.
When we apply our loss function as auxiliary loss during end-to-end training, we jointly optimize this loss function with the metric loss, as will be shown in Section~\ref{sec:method-auxiliary-loss}.
Intuitively, we want to ensure that activations are not correlated between groups. 
For a sample $\boldsymbol{x}^{(n)}$, we encourage this with the following suppression loss function
\begin{equation}
    \mathcal{L}_{\texttt{sup}_\texttt{(i,j)}}(\boldsymbol{x}^{(n)}) =  \sum_{\substack{k \in G_i, \\ l \in G_j}} (f_i(\boldsymbol{x}^{(n)})_{k} \cdot f_j(\boldsymbol{x}^{(n)})_{l})^2, 
\end{equation}
where $f_i(\boldsymbol{x}^{(n)}) = \phi(\boldsymbol{x}^{(n)})^\top \boldsymbol{W}_i$ denotes the $i$-th embedding ($1 \le i \le M$) of input image $\boldsymbol{x}^{(n)}$, $\boldsymbol{W}_i$ denotes the 
sub-matrix of $\boldsymbol{W}$ corresponding to the $i$-th embedding and $f_i(\boldsymbol{x}^{(n)})_{k}$ the $k$-th dimension of $f_i(\boldsymbol{x}^{(n)})$.
Na\"ively solving this problem, however, leads to the trivial solution $\boldsymbol{W} = \boldsymbol{0}$. To prevent this trivial solution, we add the regularization term
\begin{equation}
    \label{eq:weight-act}
    \mathcal{L}_{\texttt{weight}} = \sum_{i=1}^d (\boldsymbol{w}_i^{\top} \boldsymbol{w}_i - 1)^2 ,
\end{equation}
where $\boldsymbol{w}_i$ (with ${1 \le i \le d}$) are the row vectors of $\boldsymbol{W}$. This term forces the squared row vector norms of $\boldsymbol{W}$ to be close to $1$ and hence avoids a trivial solution.
Our final Activation Loss combines both $\mathcal{L}_{\texttt{sup}}$ and $\mathcal{L}_{\texttt{weight}}$
\begin{equation}
    \label{eq:loss-act}
    \mathcal{L}_{\texttt{act}} = \frac{1}{N} \sum_{n=1}^N \sum_{\substack{i = 1, \\ j = i + 1}}^M \mathcal{L}_{\texttt{sup}_\texttt{(i,j)}}(\boldsymbol{x}^{(n)}) + \lambda_{\texttt{w}} \cdot \mathcal{L}_{\texttt{weight}}, 
\end{equation}
where $\lambda_{\texttt{w}}$ is a regularization parameter, which we set high enough such that all row-vectors have a squared norm close to ${1 \pm 1e^{-3}}$.

\subsubsection{Adversarial Loss}
\label{sec:adversarial-loss}

\begin{figure}[t]
    \begin{center}
    \includegraphics[width=0.5\textwidth]{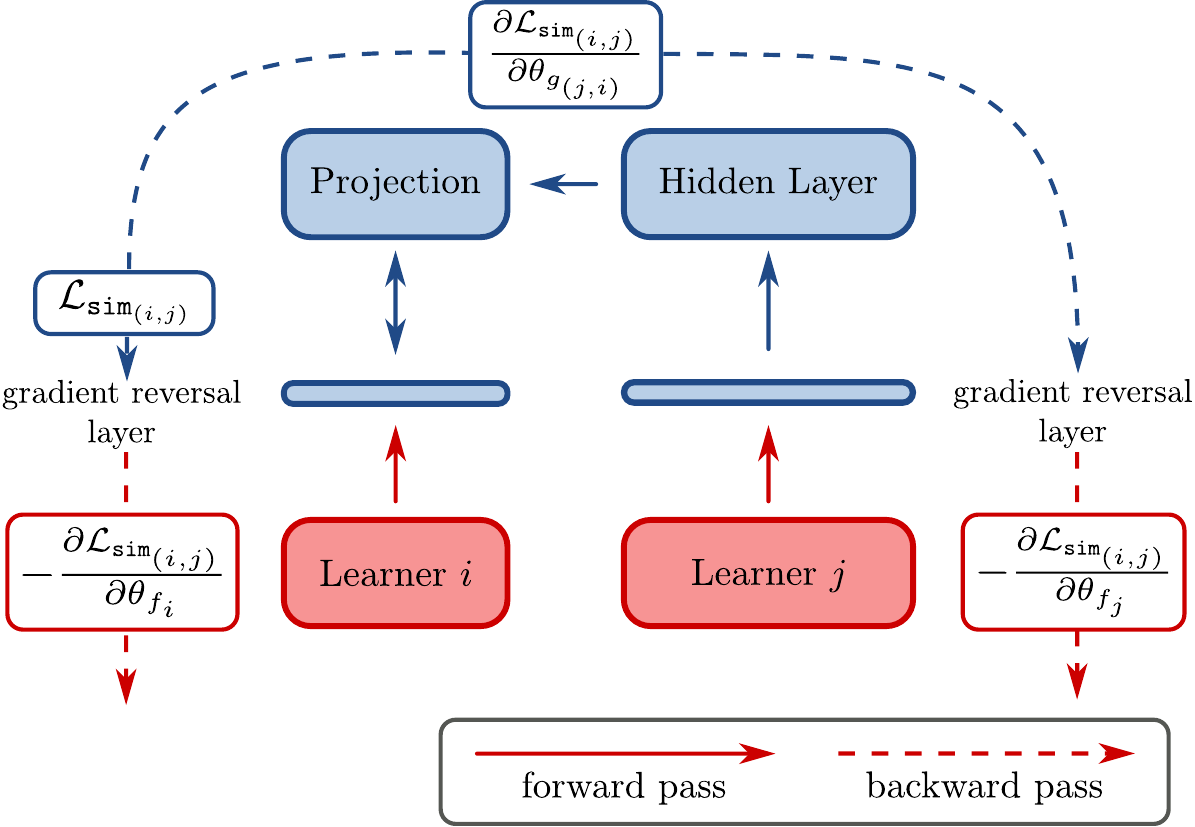}
    \end{center}
    \caption{Illustration of our adversarial regressors (blue) between learner $i$ and learner $j$ of our embedding (red). We learn a regressor which maps the vector of learner $j$ to learner $i$, maximizing the similarity of feature vectors. 
    The gradient reversal layer flips the sign of the gradients which are backpropagated to our embeddings, therefore minimizing the similarity of feature vectors. We apply these regressors during training between all pairs of our learners.}
    \label{fig:adversarial-regressors}
\end{figure}

The previous Activation Loss imposes a rather strong constraint on the embeddings, \ie for a given sample only a single embedding should be active and all other embeddings
should be close to zero.
While this improves the results, our objective is to maximize diversity between two 
feature vectors $f_i(\boldsymbol{x}) \in \mathbb{R}^{d_i}$ and $f_j(\boldsymbol{x}) \in \mathbb{R}^{d_j}$ extracted from embedding $i$ and $j$, 
where in general $d_i \neq d_j$. Rather than imposing the constraint that only a single embedding is active for a given 
training sample, we could also aim for a weaker constraint. 
We want the two vectors $f_i(\boldsymbol{x})$ and $f_j(\boldsymbol{x})$ to be different, nonetheless discriminative. Therefore, the distance
between the two vectors should be large. Unfortunately, there is no straightforward way to measure distances between two different vector spaces, since they
can \eg be of different dimensionality or be permuted.

To overcome this problem, we introduce an adversarial loss function, which we illustrate in Fig.~\ref{fig:adversarial-regressors}.
We learn regressors between pairs of embeddings 
which project $f_j(\boldsymbol{x})$ into the feature space $f_i(\boldsymbol{x})$, maximizing
the similarity between embeddings $f_i(\boldsymbol{x})$ and $f_j(\boldsymbol{x})$, by maximizing a loss function. On the other hand, our learners try to minimize this loss function \wrt these adversarial regressors and therefore maximize their diversity.
To achieve this, 
we use a reverse gradient layer~\cite{ganin2016domain} between regressors and embeddings. 
During the forward pass this layer behaves like the identity function.
However, during the backward pass, this layer flips the sign of the
gradients. As a consequence, our embedding learners minimize this loss function with respect to the regressors, \ie increasing their diversity.

More formally, let $f_m(\boldsymbol{x}) \in \mathbb{R}^{d_m}$
denote the $d_m$ dimensional embedding of the $m$-th learner. The objective of our adversarial 
regressor is to learn a function $g_{(j,i)}(\cdot): \mathbb{R}^{d_j} \mapsto \mathbb{R}^{d_i}$ from the $d_j$-dimensional embedding $j$ to the $d_i$-dimensional embedding $i$, maximizing 
similarity between vectors from embedding $j$ and $i$ via the loss
\begin{equation}
    \mathcal{L}_{\texttt{sim}_\texttt{(i,j)}}(\boldsymbol{x}^{(n)}) =  \frac{1}{d_j} \sum (f_i(\boldsymbol{x}^{(n)}) \odot g_{\left(j,i\right)}(f_j(\boldsymbol{x}^{(n)}))) ^ 2,
\end{equation}
where $\odot$ denotes the Hadamard (\ie elementwise) product.
This loss function can be made arbitrary large by scaling the weights of the regressors $g_{\left(j,i\right)}$ as well as the weights $\boldsymbol{W}$ of the embedding.
Hence, we penalize large weights $\boldsymbol{\widehat{W}}$ and biases $\boldsymbol{b}$ of $g_{\left(j,i\right)}$, and the weights $\boldsymbol{W}$ of our embedding as
\begin{align}
    \label{eq:weight-adv}
    \mathcal{L}_{\texttt{weight}} = \max(0, \boldsymbol{b}^\top \boldsymbol{b} - 1) +& \sum_i (\boldsymbol{\widehat{w}}_i^{\top} \boldsymbol{\widehat{w}}_i - 1) ^ 2 + \nonumber \\
                                    & \sum_i (\boldsymbol{w}_i^{\top} \boldsymbol{w}_i - 1)^2,
\end{align}
where $\boldsymbol{\widehat{w}}_i$ denotes the $i$-th row of the weight matrix $\boldsymbol{\widehat{W}}$ and $\boldsymbol{w}_i$ denotes the $i$-th row of the weight matrix $\boldsymbol{W}$.
We combine both terms to train the regressor with our adversarial loss
\begin{equation}
    \label{eq:adv}
    \mathcal{L}_{\texttt{adv}} = \frac{1}{N} \sum_{n=1}^N \sum_{\substack{i = 1 \\ j = i + 1}}^M -\mathcal{L}_{\texttt{sim}_\texttt{(i,j)}}(\boldsymbol{x}^{(n)}) + \lambda_{\texttt{w}} \cdot \mathcal{L}_{\texttt{weight}},
\end{equation}
where $M$ is the number of learners in our ensemble. $\lambda_{\texttt{w}}$ is a regularization parameter, which we set high enough 
so that our weight vectors have a squared norm close to ${1 \pm 1e^{-3}}$.

Backpropagating the errors of this loss function to our learners increases
their correlation and reduces their diversity. However, since we use a gradient
reversal layer between our learners and the regressors, we actually force our learners
to minimize $\mathcal{L}_{\texttt{sim}_\texttt{(i,j)}}$, consequently increasing
their diversity.  In our experiments, we use two-layer neural networks with a
\ac{ReLU} as non-linearity in the hidden layer for the regressor $g_{\left(j, i\right)}$. Further, we choose a 
hidden layer size of $512$.

We use the loss function in Eq.~\eqref{eq:adv} as auxiliary loss function
during training as shown in the following section. At test time, we simply discard the regressors.  Hence, we do
not introduce any additional parameters during test time with this adversarial
loss function. During training time, computational cost is dominated by calculating
the forward and backward pass of the convolution layers. Further, since we are only using
a gradient reversal layer as opposed to alternating updates of our adversarial network
and our base network, we can update the parameters of both networks in a single forward and backward pass.
Hence, we do not introduce significant additional computational cost during training time.

\subsection{Optimizing Diversity Loss Functions}
\label{sec:optimizing-diversity}

We present two ways to apply the previously defined loss functions to
improve our boosting based method.
In our first approach, we use one of our diversity loss functions, \ie either our Activation Loss or our Adversarial Loss, for initializing the embedding matrix $\boldsymbol{W}$. We fix all lower level \ac{CNN} parameters 
and solve an optimization problem for the embedding matrix $\boldsymbol{W}$. 
Then, we perform end-to-end training of the \ac{CNN} with this initialization and
our boosting based method (Section~\ref{sec:initialization-method}).
Our second method applies the diversity loss during training time as auxiliary loss together with our 
boosting based method (Section~\ref{sec:method-auxiliary-loss}).

\subsubsection{Initialization Method}
\label{sec:initialization-method}

During initialization we want to find an initial estimate of the embedding matrix $\boldsymbol{W}$, so that 
our learners already have low correlation with each other at the beginning of the training.
Therefore, we omit end-to-end training and instead fix all the \ac{CNN} parameters except the embedding
matrix $\boldsymbol{W}$. We minimize a loss function which encourages diversity of learners by solving 
the following optimization problem with \ac{SGD} and momentum
\begin{equation}
    \argmin_{\boldsymbol{W}} \mathcal{L_{\texttt{div}}},
\end{equation}
where $\mathcal{L_{\texttt{div}}}$ is either $\mathcal{L_{\texttt{act}}}$ (\cf~Eq.~\eqref{eq:loss-act}) if we 
use our Activation Loss or $\mathcal{L_{\texttt{adv}}}$ (\cf~Eq.~\eqref{eq:adv}) if we use our Adversarial Loss.

Compared to training
a full \ac{CNN}, solving this problem takes only seconds to a few minutes depending
on the size of the dataset. The main reason for this is that we can pre-compute all
lower level \ac{CNN} features and just optimize with respect to the last layer
(\ie the embedding matrix $\boldsymbol{W}$). As a consequence, the number of
parameters for which we are optimizing is smaller and the computational load
is lower, hence convergence is quicker.

We show the benefits of both, our Adversarial Loss and Activation Loss as initialization method in Section~\ref{sec:eval-initialization}.
Both loss functions significantly improve the accuracy of our boosting based method, as they reduce the correlation between
embeddings already from the beginning of the training.

\subsubsection{Auxiliary Loss Function}
\label{sec:method-auxiliary-loss}

When we apply the loss functions as auxiliary loss during training, we sample
the matrix $\boldsymbol{W}$ uniformly random~\cite{Glorot10} and introduce an additional
weighting parameter $\lambda_\texttt{div}$, which controls the strength of our diversity regularizer.
More formally, during training time we optimize the following loss
\begin{equation}
    \label{eq:training}
    \mathcal{L} = \mathcal{L}_{\texttt{metric}} + \lambda_{\texttt{div}} \cdot \mathcal{L}_{\texttt{div}}, 
\end{equation}
where $\mathcal{L}_{\texttt{metric}}$ is the discriminative metric loss (\eg binomial deviance, contrastive, triplet), which is minimized by our boosting based algorithm 
and $\mathcal{L}_{\texttt{div}}$ is 
our loss function which encourages diversity in our ensemble. We either use $\mathcal{L_{\texttt{act}}}$ (Eq.~\eqref{eq:loss-act}) or $\mathcal{L_{\texttt{adv}}}$ (Eq.~\eqref{eq:adv}) for $\mathcal{L_{\texttt{div}}}$, depending on whether we use our Activation Loss or our Adversarial Loss, respectively. The weighting parameter $\lambda_{\texttt{div}}$ controls the strength
of the diversity and can be set via cross-validation.
We found it necessary to backpropagate the gradients of this auxiliary loss function only to the last layer of the \ac{CNN}, \ie the embedding layer. 
The main reason for this is that setting parameters of a \ac{CNN} to $\textbf{0}$ allows a trivial optimal solution for all 
our loss functions. To prevent this collapse, we add a constraint on the weights of our network (see Eq.~\eqref{eq:weight-act} and Eq.~\eqref{eq:weight-adv}). For the embedding layer, we typically constrain 
the weights to have a squared $L_2$ norm of $1$ for
all row vectors. Adding this constraint to the hidden layers of a \ac{CNN}, however, corrupts the learned ImageNet features. 
Hence, we only backpropagate this loss to the embedding layer, which we add on top of the last hidden layer of our feature extractor.
During training time this has only a small computational overhead compared to standard backpropagation, as only the last layer is affected.

We show the benefits of using our Activation Loss and our Adversarial Loss
as auxiliary loss function in
Section~\ref{sec:eval-auxiliary-loss-function}. When applied as auxiliary loss, our Adversarial
Loss is more effective than our Activation Loss, \ie it reduces the correlation
between embeddings more without impairing their accuracy and as a result achieves
higher ensemble accuracy.

\section{Evaluation}
\label{sec:evaluation}

We first conduct a detailed ablation study on the CUB-200-2011~\cite{WahCUB_200_2011} dataset.
We follow the evaluation protocol proposed
in~\cite{oh2016deep} and use the first 100
classes ($5,864$ images) for training and the remaining $100$ classes ($5,924$
images) for testing.

For evaluation we use the Recall@$K$ metric~\cite{oh2016deep}. For each
image in the test set, we compute the feature vectors from our \ac{CNN} and
then retrieve the $K$ most similar images from the remaining test set. If one
of the $K$ retrieved images has the same label as the query image, it is a
match and increases the recall score by $1$.  The final Recall@$K$ score is the
average over all test images.

We implement our method with Tensorflow~\cite{tensorflow2015-whitepaper}. 
As network architecture, we follow previous works (\eg~\cite{oh2016deep, ustinova2016histogram}) and use a GoogLeNet\footnote{We dump the weights of the network from the Caffe~\cite{jia2014caffe} model.}~\cite{Szegedy2014} which is pre-trained
on the ImageNet dataset~\cite{ILSVRC15}.
As optimization method we use ADAM~\cite{KingmaB14} with a learning rate of $1e^{-6}$. When we use auxiliary loss functions, we can increase 
the learning rate by an order of magnitude to $1e^{-5}$ (see Section~\ref{sec:eval-auxiliary-loss-function}). 
We construct a mini-batch
by first sampling a fixed number of categories from the dataset and then sampling several
images for each of these categories. Each mini-batch consists of approximately $5$-$10$ images
per category. 

For preprocessing, we follow previous work, \eg~\cite{oh2016deep, ustinova2016histogram} and 
resize the longest axis of our images to $256$ pixels and pad the shorter axis with white pixels such that images have a size of 
$256 \times 256$ pixels. We subtract the mean from the ImageNet dataset channel-wise from the image.
During training time, we crop random $224 \times 224$ pixel patches from
the images and randomly mirror them. During test time, we use the $224 \times 224$ pixel center crop 
from an image to predict the final feature vector used for retrieval.

In the following section, we show the impact of an ensemble trained with \ac{BIER} on the strength (\ie accuracy)
and correlation of an embedding (Section~\ref{sec:eval:strength-and-correlation}). Next, we show that \ac{BIER} works with several widely used loss functions (Section~\ref{sec:eval-loss-functions}),
we analyse the impact of the number of groups in an embedding (Section~\ref{sec:eval-number-of-groups}) and the embedding size (Section~\ref{sec:eval-embedding-size}).
Then, we demonstrate the effectiveness of our diversity loss functions during initialization (Section~\ref{sec:eval-initialization}) and as 
auxiliary loss function during training (Section~\ref{sec:eval-auxiliary-loss-function}). We show the influence of our weighting parameter $\lambda_\text{div}$ (Section~\ref{sec:eval-regularization-lambda}). 
Finally, we show that our method
outperforms state-of-the-art methods on several 
datasets~\cite{krause20133d,liu2016deep,liu2016deepfashion,oh2016deep,WahCUB_200_2011} (Section~\ref{sec:eval-sota}).

\subsection{Strength and Correlation}
\label{sec:eval:strength-and-correlation}

The performance of an ensemble depends on two elements: the strength (\ie accuracy) of 
individual learners and the correlation between the learners~\cite{breiman2001random}. 
Ideally, learners of an ensemble are highly accurate and lowly correlated, 
so that they can complement each other well.

To evaluate the impact of our contributions on strength and correlation, we compare 
several models. First, we train a model with
a regular loss function with an embedding size of $512$ (\emph{Baseline}).  Next, we use a simple model averaging approach, where we
split the last embedding layer into three non-overlapping groups of size $170$,
$171$ and $171$ respectively, initialize them with our Activation Loss initialization method and optimize 
a discriminative metric loss function on each of these
groups separately (\emph{Init-170-171-171}). 
Finally, we apply our boosting based
reweighting scheme on the three groups (\emph{BIER-170-171-171}).  

As discussed in Section~\ref{sec:gradient-boosting}, we propose to use groups of 
different sizes proportional to the weighting of the online boosting algorithm, as subsequent learners have to deal with harder samples. To this end, we divide the embedding 
into differently sized groups. We assign
the first learner a size of $96$ neurons, the second learner $160$ neurons and the last 
learner $256$ neurons. Finally, we train a model with our Activation Loss initialization method (\emph{Init-96-160-256}) and 
add our boosting method (\emph{BIER-96-160-256}) on top of these learners.

As shown in Table~\ref{tbl:strength-and-correlation}, initializing
the weight matrix such that activations are independent already achieves a notable improvement over our baseline model. 
Additionally, our boosting method significantly increases the accuracy of the ensemble.
Without boosting, the
individual classifiers are highly correlated. By training successive
classifiers on reweighted samples, the classifiers focus on different
training examples leading to less correlated classifiers. 
Interestingly, the individual weak learners trained with just our Activation Loss initialization method achieve similar accuracy compared to our
boosted learners (\eg $51.94$ vs $51.47$ of \emph{Learner-1-170}), but the combination achieves a significant improvement since each group focuses
on a different part of the dataset.

\begin{table}[htbp]
\caption{Evaluation of classifier (Clf.) and feature correlation on CUB-200-2011~\cite{WahCUB_200_2011}. \textbf{Best} results are highlighted.}
\label{tbl:strength-and-correlation}
\renewcommand{\arraystretch}{1.3}

\centering

\begin{tabular}{lllll}
\hline
Method                                  & Clf. Corr. $\downarrow$ & Feature Corr. $\downarrow$ & R@1 $\uparrow$ \\
\hline
Baseline-512                            & -                & 0.1530            & 51.76  \\
\hline
Init-170-171-171                        & 0.8362           & 0.1005              & 53.73 \\ 
\hspace{2mm}  Learner-1-170      &                  &                     & 51.94 \\
\hspace{2mm}  Learner-2-171      &                  &                     & 51.99 \\
\hspace{2mm}  Learner-3-171      &                  &                     & 52.26 \\
\hline
Init-96-160-256                         & 0.9008           & 0.1197           & 53.93 \\
\hspace{2mm}  Learner-1-96       &                  &                     & 50.35 \\
\hspace{2mm}  Learner-2-160      &                  &                     & 52.60 \\
\hspace{2mm}  Learner-3-256      &                  &                     & 53.36 \\
\hline
BIER-170-171-171                        & 0.7882           & 0.0988           & 54.76 \\
\hspace{2mm}  Learner-1-170      &                  &                     & 51.47 \\
\hspace{2mm}  Learner-2-171      &                  &                     & 52.28 \\
\hspace{2mm}  Learner-3-171      &                  &                     & 52.38 \\
\hline
BIER-96-160-256                  & \textbf{0.7768}  & \textbf{0.0934}     & \textbf{55.33}  \\
\hspace{2mm}  Learner-1-96       &                  &                     & 49.95 \\
\hspace{2mm}  Learner-2-160      &                  &                     & 52.82 \\
\hspace{2mm}  Learner-3-256      &                  &                     & 54.09 \\
\hline
\end{tabular}
\end{table}

\subsection{Loss Functions}
\label{sec:eval-loss-functions}

To show that \ac{BIER} works with several loss functions such as triplet loss or contrastive loss,
we train a baseline \ac{CNN} with embedding size of $512$ and then with our boosting based method. 
For our method, we set the group size to $96$, $160$ and $256$ respectively. In Table~\ref{tbl:loss-functions} we see that 
binomial deviance, triplet loss and contrastive loss can benefit from our method.

Further, we see that our method performs best for loss functions with smooth
(\ie continuous) gradient. We hypothesize that this is due to the fact that
non-smooth loss functions convey less information in their gradient. The
gradient of the triplet and contrastive loss (for negative samples) is either $0$
or $1$, whereas the gradient of binomial deviance has continuous values
between $0$ and $1$.  

\begin{table}[htbp]
\caption{Evaluation of loss functions on CUB-200-2011~\cite{WahCUB_200_2011}. }
\label{tbl:loss-functions}

\renewcommand{\arraystretch}{1.3}
\centering
\begin{tabular}{lllll}
\hline
Method                         & Feature Corr. $\downarrow$    & R@1 $\uparrow$  \\
\hline
Triplet-512                    & 0.2122          & 50.12 \\
Triplet-96-160-256             & \textbf{0.1158}  & \textbf{53.31}  \\
\hline
Contrastive-512                & 0.1639          & 50.62            \\
Contrastive-96-160-256         & \textbf{0.1246}  & \textbf{53.8}   \\
\hline
Binomial-Deviance-512        &  0.1530            & 51.76  \\
Binomial-Deviance-96-160-256 & \textbf{0.0934}  & \textbf{55.33}  \\
\hline
\end{tabular}
\end{table}

\subsection{Number of Groups}
\label{sec:eval-number-of-groups}

We demonstrate the influence of the number of groups on our method. To this end,
we fix the embedding size to $512$ and 
run our method with $M = \left\{2, 3, 4, 5\right\}$ groups. The
group size is proportional to the final weights of our boosting
algorithm (see Section~\ref{sec:gradient-boosting}). In Table~\ref{tbl:group-sizes}
we report the correlation of the feature embedding, the R@1 score of the ensemble and the average
of the R@1 score of each individual learner.
We see that with a fixed embedding size of $512$, the optimal number of
learners for our method is $3$-$4$. For a larger number of groups the strength of individual learners 
declines and hence performance decreases. For a smaller number of groups
the individual embeddings are larger. They achieve higher individual accuracy, but are more correlated
with each other, since they benefit less from the gradient
boosting algorithm.

\begin{table}[htbp]
    \caption{Evaluation of group sizes on CUB-200-2011~\cite{WahCUB_200_2011}.}
    \label{tbl:group-sizes}
    \renewcommand{\arraystretch}{1.3}
    \centering
        \begin{tabular}{lllll}
            \hline

            \hline
            Group Sizes       &  Clf. Corr. $\downarrow$      & Avg R@1 $\uparrow$       &  R@1 $\uparrow$ \\
            \hline
            Baseline          & -                & -              & 51.76 \\
            \hline
            170-342           &   0.8252         & \textbf{53.06} & 54.66  \\
            96-160-256        &  0.7768          & 52.29          & 55.33 \\
            52-102-152-204    &  0.7091          & 50.67          & \textbf{55.62} \\
            34-68-102-138-170 &  \textbf{0.6250} & 48.5           & 54.9  \\
            \hline

        \end{tabular}
\end{table}

\subsection{Embedding Sizes}
\label{sec:eval-embedding-size}

Next, we show the effect of different embedding sizes. We train a CNN with embedding sizes of $384$, $512$, $1024$ with \ac{BIER}
and compare it to a regular \ac{CNN}. For our method, we split the embeddings into several groups according to the weights of the learners (see Section~\ref{sec:gradient-boosting}). 
We divide the $384$ sized embedding into groups  of size $64$, $128$ and $192$, respectively. For the embedding of size $512$ we 
use groups of size $96$, $160$ and $256$. Finally, for the largest embedding we use groups of 
size $50$, $96$, $148$, $196$, $242$ and $292$.

We use the binomial deviance loss function, as it consistently achieves best
results compared to triplet loss or contrastive loss (recall
Table~\ref{tbl:loss-functions}). In Table~\ref{tbl:eval-embedding-sizes} we see
that our method yields a consistent gain for a variety of different embedding sizes. For larger embedding 
sizes a larger number of groups is more beneficial. 
We found that the main reason for this is that larger embeddings are more likely to over-fit. 
Hence, it is more beneficial to train several smaller learners which complement each other better.

\begin{table}[htbp]
\caption{Evaluation of embedding size on CUB-200-2011~\cite{WahCUB_200_2011}.}
\label{tbl:eval-embedding-sizes}
\renewcommand{\arraystretch}{1.3}
\centering
\begin{tabular}{lllll}
\hline
Method                        & Feature Corr. $\downarrow$    & R@1 $\uparrow$           \\
\hline
Baseline-384                  & 0.1453           & 51.57          \\
BIER-64-128-192               & \textbf{0.0939}  & \textbf{54.66} \\
\hline
Baseline-512                  &  0.1530          & 51.76         \\
BIER-96-160-256               & \textbf{0.0934}  & \textbf{55.33}  \\
\hline
Baseline-1024                 & 0.1480           & 52.89         \\
BIER-50-96-148-196-242-292    & \textbf{0.0951}  & \textbf{55.99} \\
\hline
\end{tabular}
\end{table}

\begin{figure}[htbp]
    \begin{center}
    \includegraphics[width=0.5\textwidth]{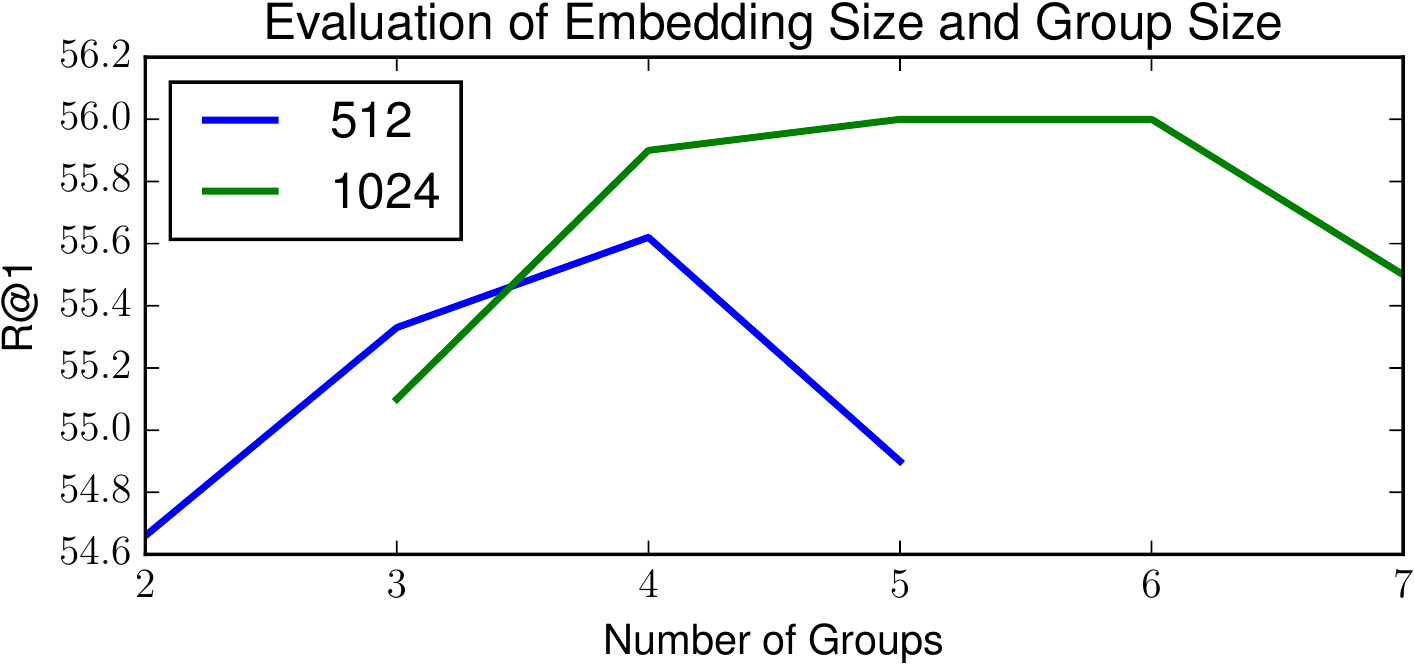}
    \end{center}
    \caption{Evaluation of different embedding sizes and group sizes on the CUB-200-2011~\cite{WahCUB_200_2011} dataset.}
    \label{fig:embedding-size-evaluation}
\end{figure}

Further, we illustrate the effect of the number of learners and the number of groups in Fig.~\ref{fig:embedding-size-evaluation}. We observe that 
with larger embedding sizes our method can use a larger number of groups. The main reason for that is that
larger embedding sizes have typically more redundancy (hence higher correlation) compared to smaller embedding sizes. Therefore, it is more beneficial to split 
a larger embedding into a larger number of groups. We set the group sizes proportional to the weight of our boosting algorithm (Section~\ref{sec:gradient-boosting}). 
For the interested reader, we also list the
corresponding group sizes in our supplementary.

\subsection{Impact of Initialization}
\label{sec:eval-initialization}

To show the effectiveness of both, our Activation Loss and Adversarial Loss for weight initialization, we compare it with 
random initialization, as proposed by Glorot~\etal~\cite{Glorot10} and an orthogonal initialization method~\cite{SaxeMG13}. 
All networks are trained with binomial deviance as loss function with our proposed boosting based reweighting scheme.
We report mean R@$1$ of the three methods.

In Table~\ref{tbl:eval-weight-init} we see that \ac{BIER} with both our initialization methods achieves better accuracy
compared to orthogonal or random initialization. This is due to the fact that with our initialization
method learners are already less correlated at the beginning of the training. This makes it easier for the boosting
algorithm to maintain diversity of our learners during training.

\begin{table}[htbp]
\caption{Evaluation of Glorot, orthogonal and our Activation Loss and Adversarial Loss initialization method on CUB-200-2011~\cite{WahCUB_200_2011}.}
\label{tbl:eval-weight-init}
\renewcommand{\arraystretch}{1.3}

\centering
\begin{tabular}{lllll}
\hline
Method                        &  R@1  \\
\hline
Glorot                        &  $54.41$ \\
Orthogonal                    &  $54.58$ \\
\hline
Activation Loss               &  $\boldsymbol{55.33}$  \\
Adversarial Loss               &  $55.04$ \\
\hline
\end{tabular}
\end{table}

\subsection{Impact of Auxiliary Loss Functions}
\label{sec:eval-auxiliary-loss-function}

To show the benefits of adding our diversity loss functions during training as auxiliary loss function,
we run several experiments on the CUB-200-2011~\cite{WahCUB_200_2011} dataset. We compare our Adversarial Loss
function to our Activation Loss function and a network which does not use an auxiliary loss function during training.
All networks are trained with the boosting based reweighting scheme and use an embedding size of $512$ with $3$ groups (\ie $96$, $160$ and $256$ learners). 
Further, we observe that we can train our networks with auxiliary loss function with an order of magnitude higher learning
rate (\ie $1e^{-5}$ instead of $1e^{-6}$), which results in significantly faster convergence times.
We report the R@1 accuracy of all our methods.

As we can see in Table~\ref{tbl:eval-adversarial-loss-function}, by including an
auxiliary loss during training we significantly improve over our previous
baseline \ac{BIER}~\cite{opitz2017bier}, which used our boosting based training but the Activation Loss 
only during initialization. By including the auxiliary
loss function during training, we can improve the stability of training,
allowing our models to be trained with larger learning rates and therefore
faster convergence. 
Training \ac{BIER}~\cite{opitz2017bier} without auxiliary loss functions and with such high learning rates
yields a significant drop in performance, since training becomes too unstable.

Finally, the Adversarial Loss
function outperforms the Activation Loss function by a significant margin. We
hypothesize this is due to the fact that the Activation Loss function
constrains the individual learners too much. The Activation Loss encourages 
the ensemble that for a given training sample,
only a single learner should be active and all other learners should be close 
to zero. In contrast to that, our Adversarial Loss minimizes similarity between 
embeddings \wrt an adversarial regressor, which tries to make two vector
spaces as similar as possible under a non-linear mapping. According to our results,
minimizing similarity is more effective for reducing correlation than suppressing entire 
vector spaces.

We also analyze the impact on strength and correlation of our auxiliary loss
functions on our ensemble. We show these results in
Table~\ref{tbl:eval-auxiliary-loss-strength-correlation}. Notably, by including
an auxiliary loss function we can significantly reduce correlation of the
feature vectors as well as the correlation between classifiers. This suggests
that our auxiliary loss functions further reduce redundancies in our embedding
and therefore improve results. Compared to the Activation Loss, our Adversarial Loss
can reduce the correlation between classifiers more effectively and achieves a
higher accuracy in terms of R@1.

The individual learners of the Adversarial Loss achieve comparable accuracy to the learners of the 
Activation Loss (\ie $51.1\%$ vs $51.3\%$, $53.8\%$ vs $53.5\%$ and $55.3\%$ vs $55.2\%$).
The Adversarial Loss, however, can significantly reduce the correlation between classifiers (\ie $0.6031$ vs $0.7310$) and features (\ie 0.0731 vs 0.0882). 
As a consequence, the individual learners are more diverse from each other and complement each other better. Therefore, our Adversarial Loss achieves a significantly better
ensemble accuracy of $57.5\%$ vs $56.5\%$.

When we use our Adversarial Loss as auxiliary loss during training, in contrast to the work of Ganin~\etal~\cite{ganin2016domain}, which uses the
gradient reversal layer for domain adaptation, we do not require a dynamic schedule
for the regularization parameter $\lambda_{\texttt{div}}$ (see Section~\ref{sec:method-auxiliary-loss}). Instead, we keep
this weighting parameter fixed. Rather than scaling back the gradients
inside the gradient reversal layer, we weight the loss function of our adversarial
network with $\lambda_{\texttt{div}}$.  As a consequence, our adversarial
auxiliary network trains slower compared to our base network, which turns out to be beneficial for the training process. We hypothesize that the main reason for this is that the
adversarial network gets too strong if we update it too fast, which in turn
degrades the performance of the base network.

\begin{table}[htbp]
    \caption{Comparison of several auxiliary loss functions on CUB-200-2011~\cite{WahCUB_200_2011}. Our adversarial loss function significantly improves accuracy over our baseline (\ac{BIER}~\cite{opitz2017bier}) and enables higher learning rates and faster convergence.}
    \label{tbl:eval-adversarial-loss-function}
    \renewcommand{\arraystretch}{1.3}
    \centering
    \begin{tabular}{llll}
        \hline
        Method & R@1 & Learning Rate & Iterations \\
        \hline
        No Auxiliary Loss & $55.3$ & $1e^{-6}$ & 50K \\
        No Auxiliary Loss & $52.3$ & $1e^{-5}$ & 15K \\
        Activation Loss  & $56.5$  & $1e^{-5}$ & 15K \\
        Adversarial Loss & $\boldsymbol{57.5}$ & $1e^{-5}$ & 15K  \\
        \hline
    \end{tabular}
\end{table}

\begin{table}[htbp]
    \caption{Impact of auxiliary loss functions on strength and correlation in the ensemble.}
    \label{tbl:eval-auxiliary-loss-strength-correlation}
    \renewcommand{\arraystretch}{1.3}
    \centering
    \begin{tabular}{lllll}
        \hline
        Method                         & Clf. Corr. $\downarrow$ & Feature Corr. $\downarrow$    & R@1 $\uparrow$            \\
        \hline
        BIER-96-160-256                & 0.7768     & 0.0934           & 55.3  \\
        \hspace{2mm}  Learner-1-96       &                  &                     & 50.0 \\
        \hspace{2mm}  Learner-2-160      &                  &                     & 52.8 \\
        \hspace{2mm}  Learner-3-256      &                  &                     & 54.1 \\
        \hline
        Activation BIER-96-160-256     & 0.7130     &  0.0882               &            56.5     \\
        \hspace{2mm}  Learner-1-96       &                  &                     & 51.3 \\
        \hspace{2mm}  Learner-2-160      &                  &                     & 53.5 \\
        \hspace{2mm}  Learner-3-256      &                  &                     & 55.2 \\
        \hline
        Adversarial BIER-96-160-256    & \textbf{0.6031} & \textbf{0.0731}           & \textbf{57.5}  \\
        \hspace{2mm} Learner-1-96                      &            &                  & 51.1 \\
        \hspace{2mm} Learner-2-160                      &            &                  & 53.8 \\
        \hspace{2mm} Learner-3-256                      &            &                  & 55.3 \\
        \hline
    \end{tabular}
\end{table}

\subsection{Evaluation of the Regularization Parameter}
\label{sec:eval-regularization-lambda}

When we add our diversity loss functions during training time we introduce an 
additional parameter $\lambda_{\texttt{div}}$ (recall Section~\ref{sec:method-auxiliary-loss}). To demonstrate its effect, we train several models on the 
CUB-200-2011 dataset~\cite{WahCUB_200_2011} with a learning rate of $1e^{-5}$ and vary $\lambda_{\texttt{div}}$.

\begin{figure}[htbp]
    \begin{center}
        \includegraphics[width=0.5\textwidth]{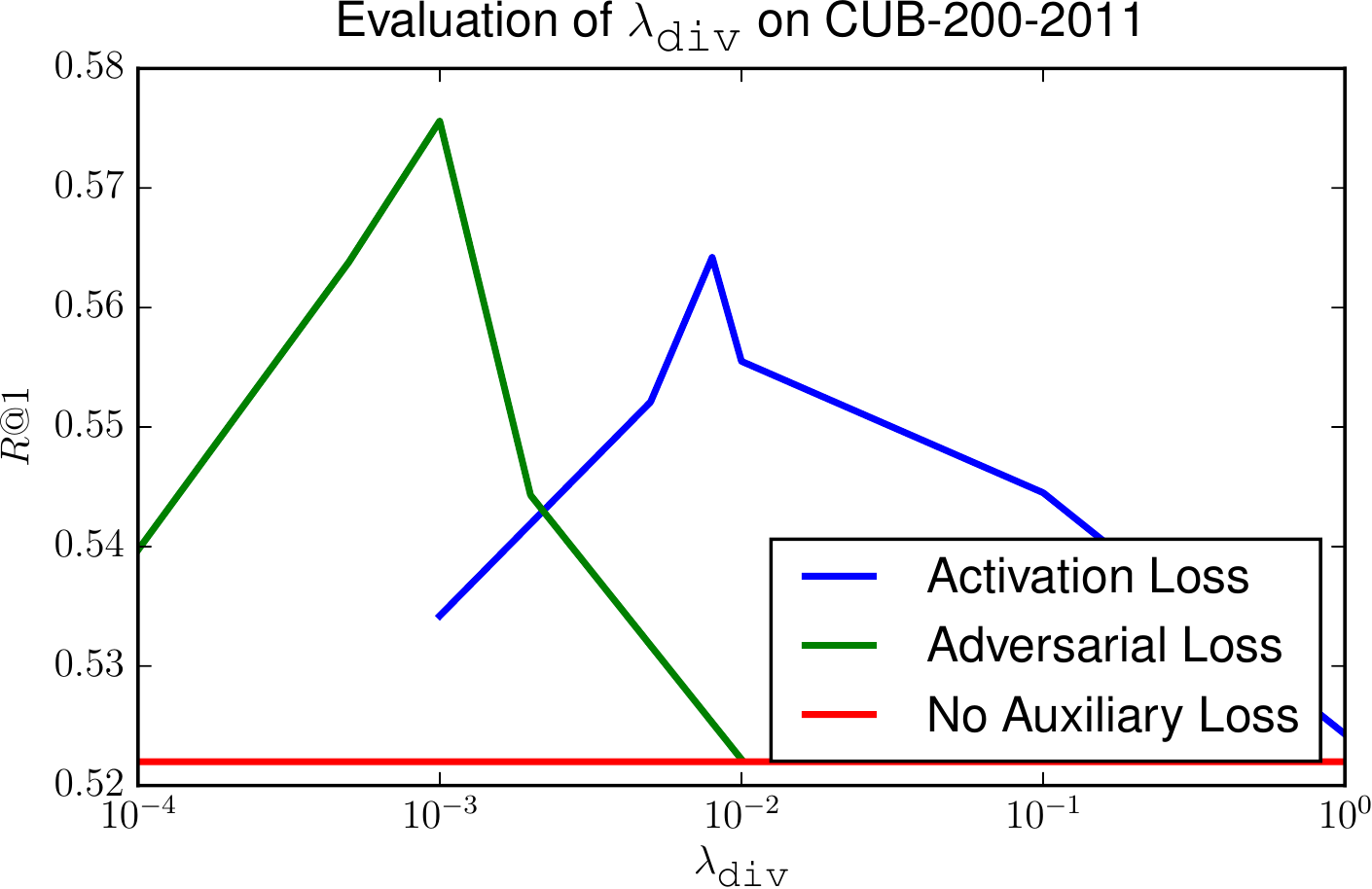}
    \end{center}
    \caption{Evaluation of $\lambda_{\texttt{div}}$ on CUB-200-2011~\cite{WahCUB_200_2011}.}
    \label{fig:eval_lambda_div_cub}
\end{figure}

In Fig.~\ref{fig:eval_lambda_div_cub} we see that for our Adversarial Loss $\lambda_{\texttt{div}}$ peaks around $1e^{-3}$, whereas
for our Activation Loss $\lambda_{\texttt{div}}$ peaks around $1e^{-2}$. Further, our Adversarial Loss significantly outperforms our 
Activation Loss by about 1\% R@1. Finally, applying any of our loss functions as auxiliary loss function with a learning rate of $1e^{-5}$ 
significantly improves R@1 compared to networks without an auxiliary loss function trained with the same learning rate.
Therefore, by integrating any of the two auxiliary loss function, we can improve the training stability of \ac{BIER} at higher learning rates.

\subsection{Comparison with the State-of-the-Art}
\label{sec:eval-sota}

\begin{table*}[!htbp]
    \caption{Comparison with the state-of-the-art on the CUB-200-2011~\cite{WahCUB_200_2011} and Cars-196~\cite{krause20133d} dataset. \textbf{Best} results are highlighted.}
    \label{tbl:sota-cub-200-2011-cars-196}
    \renewcommand{\arraystretch}{1.3}
    \centering
    \begin{tabular}{l|cccccc|cccccc}
        \hline
        & \multicolumn{6}{c}{CUB-200-2011} & \multicolumn{6}{c}{Cars-196} \\
        \hline
        R@K                                            & 1     & 2     & 4    & 8    & 16   & 32                                       & 1 & 2 & 4 & 8 & 16 & 32  \\
        \hline
        Contrastive~\cite{oh2016deep}                  & 26.4  & 37.7  & 49.8 & 62.3 & 76.4 & 85.3                                     & 21.7 & 32.3 & 46.1 & 58.9 & 72.2 & 83.4 \\
        Triplet~\cite{oh2016deep}                      & 36.1  & 48.6  & 59.3 & 70.0 & 80.2 & 88.4                                     & 39.1  & 50.4  & 63.3 & 74.5 & 84.1 & 89.8 \\
        LiftedStruct~\cite{oh2016deep}                 & 47.2  & 58.9  & 70.2 & 80.2 & 89.3 & 93.2                                     & 49.0 & 60.3 & 72.1 & 81.5 & 89.2 & 92.8 \\
        Binomial Deviance~\cite{ustinova2016histogram} & 52.8  & 64.4  & 74.7 & 83.9 & 90.4 & 94.3                                     & - & -  & - & - & - & - \\
        Histogram Loss~\cite{ustinova2016histogram}    & 50.3  & 61.9  & 72.6 & 82.4 & 88.8 & 93.7                                     & - & - & - & - & - & - \\
        N-Pair-Loss~\cite{sohn2016improved}            & 51.0  & 63.3  & 74.3 & 83.2 & -    & -                                        & 71.1 & 79.7 & 86.5 & 91.6 & -    & -   \\
        Clustering~\cite{song2017cvpr}                 & 48.2  & 61.4  & 71.8 & 81.9   & -    & -                                      & 58.1 & 70.6 & 80.3 & 87.8 & -    & -   \\
        Proxy NCA~\cite{movshovitz-attias2017iccv}     & 49.2  & 61.9  & 67.9 & 72.4   & -    & -                                      & 73.2 & 82.4 & 86.4 & 87.8 & -    & -   \\
        Smart Mining~\cite{harwood2017iccv}            & 49.8  & 62.3 & 74.1  & 83.3  & -    & -                                       & 64.7 & 76.2 & 84.2 & 90.2 & -    & -   \\
        HDC~\cite{yuan2016hard}                        & 53.6  & 65.7  & 77.0 & 85.6 & 91.5 & \textbf{95.5}          & 73.7  & 83.2  & 89.5 & 93.8 & 96.7 & 98.4 \\
        Angular Loss~\cite{wang2017iccv}               & 54.7  & 66.3  & 76.0 & 83.9 & -    & -                                        & 71.4 & 81.4 & 87.5 & 92.1 & -    & -   \\
        \hline
        Ours Baseline                                  & 51.8 & 63.8 & 74.1 & 83.1 & 90.0 & 94.8                                       & 73.6 & 82.6 & 89.0 & 93.5 & 96.4 & 98.2 \\
        BIER Learner-3~\cite{opitz2017bier}            & 54.1 & 66.1 & 76.5 & 84.7 & 91.2 & 95.3                                       & 76.5 & 84.9 & 90.9 & 94.9 & 97.6 & 98.7 \\
        \textbf{BIER}~\cite{opitz2017bier}             & 55.3 & 67.2 & 76.9 & 85.1 & 91.7 & \textbf{95.5}                              & 78.0 & 85.8 & 91.1 & 95.1 & 97.3 & \textbf{98.7} \\
        \hline
        A-BIER Learner-3                     & 55.3 & 67.0 & 76.8 & 86.0 & 91.1 & 95.3                                       & 80.6 & 88.2 & 92.3 & 95.8 & 97.6 & 98.6 \\
        \textbf{A-BIER}                      & \textbf{57.5} & \textbf{68.7} & \textbf{78.3} & \textbf{86.2} & \textbf{91.9} & \textbf{95.5} & \textbf{82.0} & \textbf{89.0} & \textbf{93.2} & \textbf{96.1} & \textbf{97.8} & \textbf{98.7} \\
        \hline
    \end{tabular}
\end{table*}
\begin{table*}[!htbp]
    \caption{Comparison with the state-of-the-art on the cropped versions of the CUB-200-2011~\cite{WahCUB_200_2011} and Cars-196~\cite{krause20133d} dataset.}
    \label{tbl:sota-cub-200-2011-cars-196-cropped}
    \renewcommand{\arraystretch}{1.3}
    \centering
    \begin{tabular}{l|cccccc|cccccc}
        \hline
        & \multicolumn{6}{c}{CUB-200-2011} & \multicolumn{6}{c}{Cars-196} \\
        \hline
        R@K                                            & 1     & 2     & 4    & 8    & 16   & 32                                       & 1 & 2 & 4 & 8 & 16 & 32  \\
        \hline

        PDDM + Triplet \cite{huang2016local}           & 50.9 & 62.1 & 73.2 & 82.5 & 91.1 & 94.4                                       & 46.4 & 58.2 & 70.3 & 80.1 & 88.6 & 92.6  \\
        PDDM + Quadruplet \cite{huang2016local}        & 58.3 & 69.2 & 79.0 & 88.4 & 93.1 & 95.7                                       & 57.4 & 68.6 & 80.1 & 89.4 & 92.3 & 94.9  \\
        HDC~\cite{yuan2016hard}       & 60.7 & 72.4 & 81.9 & 89.2 & 93.7 & 96.8                                                        & 83.8 & 89.8 & 93.6 & 96.2 & 97.8 & 98.9 \\
        Margin~\cite{wu2017iccv}                       & 63.9  & 75.3 & \textbf{84.4} & \textbf{90.6} & \textbf{94.8} & -              & 86.9 & 92.7 & 95.6 & 97.6 & 98.7 & -    \\
        \hline
        Ours Baseline                                  &  58.9 & 70.1 & 79.8 & 87.6 & 92.6 & 96.0                                      & 82.6 & 88.8 & 93.1 & 96.1 & 97.5 & 98.7 \\
        BIER Learner-3~\cite{opitz2017bier}            &  62.8 & 73.5 & 81.9 & 89.0 & 93.7 & 96.7                                      & 85.8 & 91.7 & 94.8 & 97.2 & 98.4 & 99.2 \\
        \textbf{BIER}~\cite{opitz2017bier}             &  63.7 & 74.0 & 82.5 & 89.3 & 93.8 & 96.8                                      & 87.2 & 92.2 & 95.3 & 97.4 & 98.5 & 99.3 \\
        \hline
        A-BIER Learner-3                     & 64.0 & 74.3 & 83.1 & 89.2 & 94.1 & 96.9                                       & 88.5 & 93.2 & 98.9 & 97.7 & 98.5 & 99.2 \\
        \textbf{A-BIER}                      &  \textbf{65.5} & \textbf{75.8} & 83.9 & 90.2 & 94.2 & \textbf{97.1}           & \textbf{90.3} & \textbf{94.1} & \textbf{96.8} & \textbf{97.9} &  \textbf{98.9} & \textbf{99.4} \\
        \hline
    \end{tabular}
\end{table*}
\begin{table*}[!htbp]
    \begin{minipage}[t]{0.5\textwidth}
    \renewcommand{\arraystretch}{1.3}
    \caption{Comparison with the state-of-the-art on the Stanford Online Products~\cite{oh2016deep} dataset.}
    \label{tbl:sota-products}
    \centering
    \begin{tabular}{l|ccccccc}
        \hline
        R@K                                            & 1     & 10     & 100    & 1000           \\
        \hline
        Contrastive \cite{oh2016deep}                  & 42.0 & 58.2 & 73.8 & 89.1      \\
        Triplet \cite{oh2016deep}                      & 42.1  & 63.5  & 82.5 & 94.8    \\
        LiftedStruct \cite{oh2016deep}                 & 62.1 & 79.8 & 91.3 & 97.4      \\
        Binomial Deviance \cite{ustinova2016histogram} & 65.5  & 82.3  & 92.3 & 97.6    \\
        Histogram Loss \cite{ustinova2016histogram}    & 63.9  & 81.7  & 92.2 & 97.7    \\
        N-Pair-Loss \cite{sohn2016improved}            & 67.7 & 83.8  & 93.0 & 97.8  \\
        Clustering~\cite{song2017cvpr}            & 67.0 & 83.7 & 93.2 & - \\
        HDC~\cite{yuan2016hard}          & 69.5 & 84.4 & 92.8 & 97.7     \\
        Angular Loss~\cite{wang2017iccv} & 70.9 & 85.0 & 93.5 & \textbf{98.0} \\
        Margin~\cite{wu2017iccv}         & 72.7 & 86.2 & 93.8 & \textbf{98.0} \\
        Proxy NCA~\cite{movshovitz-attias2017iccv} & 73.7 & - & - & - \\
        \hline
        Ours Baseline                                  & 66.2 & 82.3 & 91.9 & 97.4 \\
        BIER Learner-3~\cite{opitz2017bier}            & 72.5 & 86.3 & 93.9 & 97.9 \\
        \textbf{BIER}~\cite{opitz2017bier}             & 72.7 & 86.5 & \textbf{94.0} & \textbf{98.0}   \\
        \hline
        A-BIER Learner-3                     & 74.0 & 86.8 & 93.9   &  97.8 \\
        \textbf{A-BIER}                      & \textbf{74.2} & \textbf{86.9} & \textbf{94.0} & 97.8          \\
        \hline
    \end{tabular}
    \end{minipage}
    \hfill
    \begin{minipage}[t]{0.5\textwidth}
    \centering
    \caption{Comparison with the state-of-the-art on the In-Shop Clothes Retrieval~\cite{liu2016deepfashion} dataset.}
    \label{tbl:sota-clothes}
    \renewcommand{\arraystretch}{1.3}
    \begin{tabular}{l|cccccc}
    \hline
    R@K             & 1 & 10 & 20 & 30 & 40 & 50 \\
    \hline
    FasionNet + Joints \cite{liu2016deepfashion}   & 41.0 & 64.0 & 68.0 & 71.0 & 73.0 & 73.5 \\
    FasionNet + Poselets \cite{liu2016deepfashion} & 42.0 & 65.0 & 70.0 & 72.0 & 72.0 & 75.0 \\
    FasionNet \cite{liu2016deepfashion}            & 53.0 & 73.0 & 76.0 & 77.0 & 79.0 & 80.0 \\
    HDC~\cite{yuan2016hard}                        & 62.1 & 84.9 & 89.0 & 91.2 & 92.3 & 93.1 \\
    \hline
    Ours Baseline                                  & 70.6 & 90.5 & 93.4 & 94.7 & 95.5 & 96.1 \\
    BIER Learner-3~\cite{opitz2017bier}            & 76.4 & 92.7 & 95.0 & 96.1 & 96.6 & 97.0 \\
    \textbf{BIER}~\cite{opitz2017bier}             & 76.9 & 92.8 & 95.2 & 96.2 & 96.7 & 97.1 \\
    \hline
    A-BIER Learner-3                     & 82.8 & 95.0     & 96.8              & 97.4              &  97.7             & 98.0          \\
    \textbf{A-BIER}                      & \textbf{83.1} & \textbf{95.1}          & \textbf{96.9}          & \textbf{97.5}          & \textbf{97.8}          & \textbf{98.0}          \\
    \hline
    
    \end{tabular}
    \end{minipage}
\end{table*}

We show the robustness of our method by comparing it with the state-of-the-art on the 
CUB-200-2011~\cite{WahCUB_200_2011}, Cars-196~\cite{krause20133d}, 
Stanford Online Product~\cite{oh2016deep}, In-Shop Clothes Retrieval~\cite{liu2016deepfashion} and VehicleID~\cite{liu2016deep} datasets. 

CUB-200-2011 consists of $11,788$ images of $200$ bird categories. 
The Cars-196 dataset contains $16,185$ images of $196$
cars classes. The Stanford Online Product dataset consists of $120,053$ images
with $22,634$ classes crawled from Ebay. Classes are hierarchically grouped into $12$ coarse categories (\eg cup, bicycle, \etc).
The In-Shop Clothes Retrieval dataset consists of $54,642$ images with $11,735$ clothing classes.
VehicleID consists of $221,763$ images with $26,267$ vehicles.

For training on CUB-200-2011, Cars-196 and Stanford Online Products, we follow the evaluation protocol proposed
in~\cite{oh2016deep}. For the CUB-200-2011 dataset, we use the first 100
classes ($5,864$ images) for training and the remaining $100$ classes ($5,924$
images) for testing.  We further use the first $98$ classes of the Cars-196
dataset for training ($8,054$ images) and  the remaining $98$ classes for
testing ($8,131$ images).  For the Stanford Online Products dataset we use the
same train/test split as~\cite{oh2016deep}, \ie we use $59,551$
images of $11,318$ classes for training and $60,502$ images of $11,316$ classes
for testing.
For the In-Shop Clothes Retrieval dataset, we use the predefined $25,882$ training images of $3,997$ classes for training.
The test set is partitioned into a query set ($14,218$ images of $3,985$ classes) and a gallery set ($12,612$ images of $3,985$ classes).
When evaluating on VehicleID, we use the predefined $110,178$ images of $13,134$ vehicles for training and the predefined
test sets (Small, Medium, Large) for testing~\cite{liu2016deep}.

We fix all our parameters and train \ac{BIER} with the binomial deviance loss
function and an embedding size of $512$ and group size of $3$ (\ie we use
groups of size $96$, $160$, $256$). For the CUB-200-2011 and Cars-196 dataset
we follow previous work, \eg~\cite{oh2016deep}, and report our results in terms
of Recall@$K$, ${K \in \left\{1, 2, 4, 8, 16, 32\right\}}$. For Stanford
Online Products we also stick to previous evaluation
protocols~\cite{oh2016deep} and report Recall@$K$, $K \in \left\{1, 10, 100,
1000 \right\}$, for the In-Shop Clothes Retrieval dataset we compare with $K
\in \left\{1, 10, 20, 30, 40, 50\right\}$ and for VehicleID we evaluate with $K
\in \left\{1, 5\right\}$. We also report the results for the last learner in
our ensemble (\emph{BIER Learner-3}), as it was trained on the most difficult
examples.
Further, we also show the benefits of using our adversarial loss function during training time 
in combination with \ac{BIER} (\emph{A-BIER}) on all datasets and also report the last learner 
in this ensemble (\emph{A-\ac{BIER} Learner-3}).

Results and baselines are shown in
Tables~\ref{tbl:sota-cub-200-2011-cars-196},~\ref{tbl:sota-cub-200-2011-cars-196-cropped},~\ref{tbl:sota-products},~\ref{tbl:sota-clothes}
and~\ref{tbl:sota-vehicleid}.  Our method in combination with a simple loss
function operating on pairs is able to outperform or achieve comparable
performance to state-of-the-art methods relying on higher order
tuples~\cite{sohn2016improved, oh2016deep},
histograms~\cite{ustinova2016histogram}, novel loss
functions~\cite{wang2017iccv, movshovitz-attias2017iccv} or hard sample mining
strategies~\cite{yuan2016hard, harwood2017iccv, wu2017iccv}. We consistently
improve our strong baseline method by a large margin at R@$1$ on all datasets,
which demonstrates the robustness of our approach. Further, by using our
adversarial loss function during training (\emph{A-\ac{BIER}}), we significantly improve over
\ac{BIER}~\cite{opitz2017bier} and outperform state-of-the-art methods. On
CUB-200-2011 and Cars-196 we can improve over the state-of-the-art
significantly by about 2-4\% at R@1. The Stanford Online Products, the In-Shop
Clothes Retrieval and VehicleID datasets are more challenging since there are
only few ($\approx 5$) images per class.  On these datasets our auxiliary adversarial loss achieves a
notable improvement over \ac{BIER} of 1.5\%, 6.1\% and 3-6\%, respectively. \emph{A-\ac{BIER}} outperforms state-of-the-art methods on all datasets. 
Notably, even the last
learner in our adversarial ensemble (\emph{A-BIER Learner-3}), evaluated on its own, already outperforms the
state-of-the-art on most of the datasets.

\begin{table}[!h]
\caption{Comparison with the state-of-the-art on VehicleID~\cite{liu2016deep}.}
\label{tbl:sota-vehicleid}
\renewcommand{\arraystretch}{1.3}
\centering
\begin{tabular}{l|cc|cc|cc}
\hline
                        & \multicolumn{2}{c}{Small} & \multicolumn{2}{c}{Medium} & \multicolumn{2}{c}{Large} \\
\hline
R@K                               & 1    & 5    & 1    & 5    & 1    & 5 \\
\hline
Mixed Diff+CCL~\cite{liu2016deep} & 49.0 & 73.5 & 42.8 & 66.8 & 38.2 & 61.6 \\
GS-TRS loss~\cite{bai2017intraclassvar} & 75.0 & 83.0 & 74.1 & 82.6 & 73.2 & 81.9 \\
\hline
Ours Baseline                       & 78.0          & 87.5          & 73.0          & 84.7           & 67.9          & 82.4 \\
BIER Learner-3~\cite{opitz2017bier} & 82.6 & 90.5          & 79.3 & 88.0           & 75.5          & 86.0 \\
\textbf{BIER}~\cite{opitz2017bier}  & 82.6 & 90.6 & 79.3 & 88.3  & 76.0 & 86.4 \\
\hline
A-BIER Learner-3        & 86.0          & 92.7          & 83.2          & 88.6 & 81.5           &  88.6              \\
\textbf{A-BIER}         & \textbf{86.3} & \textbf{92.7} & \textbf{83.3} & \textbf{88.7} & \textbf{81.9}  &  \textbf{88.7}              \\
\hline

\end{tabular}
\end{table}

\section{Conclusion}
\label{sec:conclusion}

In this work we cast training an ensemble of metric \acp{CNN} with a shared
feature representation as online gradient boosting problem.  We further
introduce two loss functions which encourage diversity in our ensemble. We apply these loss functions either during initialization or as auxiliary loss function during training.
In our experiments we show that our loss functions increase
diversity among our learners and, as a consequence, significantly increase
accuracy of our ensemble. Further, we show that our novel Adversarial
Loss function outperforms our previous Activation Loss function. This is because our 
Adversarial Loss increases the diversity in our ensemble more effectively. Consequently, the ensemble accuracy is higher for networks trained 
with our Adversarial Loss.
Our method does not
introduce any additional parameters during test time and has only negligible
additional computational cost, both, during training and test time.  Our extensive experiments show
that \ac{BIER} significantly reduces correlation on the last hidden layer of a
\ac{CNN} and works with  several different loss functions. Finally, by training
with our auxiliary loss function Adversarial \ac{BIER} outperforms
state-of-the-art methods on the Stanford Online Products~\cite{oh2016deep},
CUB-200-2011~\cite{WahCUB_200_2011}, Cars-196~\cite{krause20133d}, In-Shop
Clothes Retrieval~\cite{liu2016deepfashion} and VehicleID~\cite{liu2016deep}
datasets.

\ifCLASSOPTIONcompsoc
  \section*{Acknowledgments} 

\else
  \section*{Acknowledgment}
\fi

This project was supported by the Austrian Research Promotion Agency (FFG) projects MANGO (836488) and Darknet (858591). We gratefully acknowledge the support of NVIDIA Corporation with the donation of GPUs used for this research.

\ifCLASSOPTIONcaptionsoff
  \newpage
\fi

\bibliographystyle{IEEEtran}
\bibliography{abbreviation_short,egbib}

\vspace{-1.0cm}
\begin{IEEEbiography}[{\includegraphics[width=1in,height=1.125in,clip,keepaspectratio]{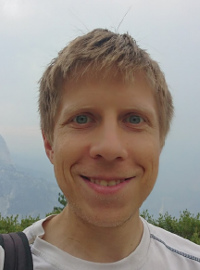}}]{Michael Opitz}
received a MSc in Visual Computing from Vienna University of Technology, Austria, in 2013. He is currently working towards his PhD degree in the Institute of Computer Graphics and Vision, Graz University of Technology, Austria. His research interests include computer vision and machine learning with a focus on metric learning and object detection.
\end{IEEEbiography}

\vspace{-1.0cm}
\begin{IEEEbiography}[{\includegraphics[width=1in,height=1.125in,clip,keepaspectratio]{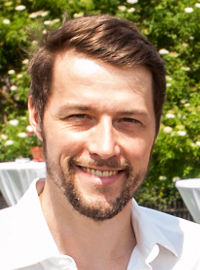}}]{Georg Waltner}
received a MSc in Telematics in 2014 from Graz University of Technology, Austria. He is currently working towards his PhD degree in the Institute of Computer Graphics and Vision, Graz University of Technology, Austria. His research interest lies in classification of food items with respect to finding optimal embedding spaces and inserting new possibly unseen categories. 
\end{IEEEbiography}

\vspace{-1.0cm}
\begin{IEEEbiography}[{\includegraphics[width=1in,height=1.125in,clip,keepaspectratio]{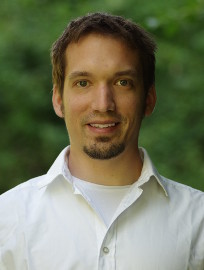}}]{Horst Possegger}
    is currently a Computer Science PhD student at the Institute for Computer Graphics and Vision Graz University of Technology, Austria. He received the BSc and MSc degrees in Software Development and Business Management from Graz University of Technology in 2011 and 2013, respectively. His research interests include visual object tracking and detection, human behaviour analysis, and video analysis in general.
\end{IEEEbiography}

\vspace{-1.0cm}
\begin{IEEEbiography}[{\includegraphics[width=1in,height=1.125in,clip,keepaspectratio]{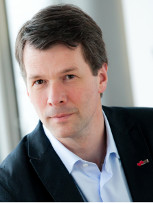}}]{Horst Bischof}
Horst Bischof received his MSc and Ph.D. degree in computer science from the Vienna University of Technology in 1990 and 1993. In 1998, he got his Habilitation (venia docendi) for applied computer science. Currently, he is Vice Rector for Research at Graz University of Technology and Professor at the Institute for Computer Graphics and Vision. 
He has published more than 650 peer reviewed scientific papers.

Horst Bischof is member of the European academy of sciences and has received several awards (20), among them the 29th Pattern Recognition award in 2002, the main price of the German Association for Pattern Recognition (DAGM) in 2007 and 2012, the best scientific paper award at the BMCV~2007, the BMVC best demo award 2012 and the best scientific paper awards at the ICPR~2008, ICPR~2010, PCV~2010, AAPR~2010 and ACCV~2012.
\end{IEEEbiography}

\appendices

\section{Overview}
\label{sec:introduction}

In this appendix to our main manuscript we provide further insights into \ac{BIER}.  First, in
Appendix~\ref{sec:algorithm} we describe our method for loss functions operating
on triplets. 
Next, in Appendix~\ref{sec:embedding-sizes-groups} we list the group sizes we used in our experiments (Section~4.4 of the main manuscript).
Further, in Appendix~\ref{sec:impact-matrix-initialization} we summarize the effect of
our boosting based training approach, our initialization approaches and our auxiliary loss functions. We
provide an experiment evaluating the impact of end-to-end training in
Appendix~\ref{sec:end-to-end-training}. Further, in
Appendix~\ref{sec:general-applicability} we demonstrate that our method is
also applicable to generic image classification problems.  
Finally, we show a qualitative comparison of the different embeddings 
in our ensemble in Appendix~\ref{sec:qualitative-comparison-of-embeddings} 
and conclude with qualitative results in Appendix~\ref{sec:qualitative-results}.

\section{BIER for Triplets}
\label{sec:algorithm}

For loss functions operating on triplets of samples, we illustrate our training method in Algorithm~\ref{algo:triplets}. 
In contrast to our tuple based algorithm, we sample triplets $\boldsymbol{x}^{(1)}$, $\boldsymbol{x}^{(2)}$ 
and $\boldsymbol{x}^{(3)}$ which satisfy the constraint that the first pair $(\boldsymbol{x}^{(1)}$, $\boldsymbol{x}^{(2)})$ 
is a positive pair (\ie $y^{(1),(2)} = 1$) and the 
second pair $(\boldsymbol{x}^{(1)}$, $\boldsymbol{x}^{(3)})$ is a negative pair (\ie $y^{(1),(3)} = 0$).
We accumulate the positive and negative similarity scores separately in the forward pass. 
In the backward pass we reweight the training set for each learner $m$ according to the negative gradient $\ell'$
at the ensemble predictions of both image pairs up to stage $m-1$.

\setlength{\algomargin}{0.0cm}
\begin{algorithm}[htbp]
    Let $\eta_m = \frac{2}{m+1}$,  for $m = 1,2,\ldots,M$, \\
    $M$ = number of learners,
    $I$ = number of iterations \\
    \For {$n = 1$ \KwTo $I$ } {
        \tcc{Forward pass} 
        Sample triplet $(\boldsymbol{x}^{(1)}_n$, $\boldsymbol{x}^{(2)}_n$, $\boldsymbol{x}^{(3)}_n)$, \\ 
            \quad s.t. $y^{(1),(2)} = 1$ and $y^{(1),(3)} = 0$.  \\
        $s_n^{0^+} := 0$ \\
        $s_n^{0^-} := 0$ \\
        \For {m = 1 \KwTo M } {
            \hspace{-0.2cm} $s_n^{m^+} := (1 - \eta_m) s_n^{{m-1}^+} + \eta_m s(f_m(\boldsymbol{x}^{(1)}_n), f_m(\boldsymbol{x}^{(2)}_n))$

            \hspace{-0.2cm} $s_n^{m^-} := (1 - \eta_m) s_n^{{m-1}^-} + \eta_m s(f_m(\boldsymbol{x}^{(1)}_n), f_m(\boldsymbol{x}^{(3)}_n))$
        }
        Predict $s_n^+$ = $s_n^{M^+}$ \\
        Predict $s_n^-$ = $s_n^{M^-}$ \\
        \vspace{2.5mm}
        \tcc{Backward pass} 
        $w_n^1 := 1$ \\
        \For {$m = 1$ \KwTo $M$ } {
            $s^{(1),(2)}_m := s(f_m(\boldsymbol{x}^{(1)}_n), f_m(\boldsymbol{x}^{(2)}_n)$ \\
            $s^{(1),(3)}_m := s(f_m(\boldsymbol{x}^{(1)}_n), f_m(\boldsymbol{x}^{(3)}_n)$ \\
            Backprop $w_n^m \ell(s^{(1),(2)}_m, s^{(1),(3)}_m)$ \\
            $w_n^{m+1} := -\ell'(s_n^{m^+}, s_n^{m^-})$
        }
    }
    
    \label{algo:triplets}
    \caption{Online gradient boosting algorithm for our \ac{CNN} using triplet based loss functions.}
\end{algorithm}

\section{List of Group Sizes}
\label{sec:embedding-sizes-groups}

In our second experiment in Section~4.4 of the main manuscript we evaluate the impact of the number of groups with 
embeddings of dimensionality $512$ and $1024$.
For the sake of clarity, we list in Table~\ref{tbl:group-sizes} the dimensionality of all groups.

\begin{table}[htbp]
    \caption{Group sizes used in our experiments.}
    \label{tbl:group-sizes}
    \renewcommand{\arraystretch}{1.3}
    \centering
    \begin{tabular}{lcl}
        \hline
        Embedding & Number of Groups & Groups \\
        \hline
        512            & 2 & 170-342 \\
        512            & 3 & 96-160-256 \\
        512            & 4 & 52-102-152-204 \\
        512            & 5 & 34-68-102-138-170 \\
        \hline
        1024           & 3 & 170-342-512 \\
        1024           & 4 & 102-204-308-410 \\
        1024           & 5 & 68-136-204-274-342 \\
        1024           & 6 & 50-96-148-196-242-292 \\
        1024           & 7 & 36-74-110-148-182-218-256 \\
        \hline
    \end{tabular}
\end{table}

\section{Summary of Improvements}
\label{sec:impact-matrix-initialization}

For the sake of clarity, we summarize our contributions on the CUB-200-2011 dataset~\cite{WahCUB_200_2011} in Table~\ref{tbl:boosting-impact}.
Our initialization method, our boosting
based training method and our auxiliary loss functions improve the final R@1 score of the model.

\begin{table}[htbp]
    \caption{Summary of the impact of our initialization method and boosting on the CUB-200-2011 dataset.}
    \label{tbl:boosting-impact}
    \renewcommand{\arraystretch}{1.3}
    \centering
    \begin{tabular}{ll}
        \hline
        Method & R@1 \\
        \hline
        Baseline & 51.76 \\
        Activation Loss initialization & 53.73 \\
        Boosting with random initialization & 54.41 \\
        Boosting with Activation Loss initialization & 55.33 \\
        \hline
        Boosting with auxiliary Activation Loss & 56.5 \\
        Boosting with auxiliary Adversarial Loss & 57.5 \\
        \hline
    \end{tabular}

\end{table}

\section{Evaluation of End-to-End Training}
\label{sec:end-to-end-training}

To show the benefits of end-to-end training with our method we apply our online
boosting approach to a finetuned network and fix all hidden layers in the
network (denoted as \emph{Stagewise training}). We compare the results against end-to-end training with \ac{BIER} with no auxiliary loss function during training time and summarize the
results in Table~\ref{tbl:boosting-end-to-end}. End-to-end training significantly improves the final R@1 score, since 
weights of lower layers benefit from the increased diversity of the ensemble.

\begin{table}[htbp]
    \caption{Influence of end-to-end training on the CUB-200-2011 dataset.}
    \label{tbl:boosting-end-to-end}
    \renewcommand{\arraystretch}{1.3}
    \centering
    \begin{tabular}{ll}
    \hline
    Method & R@1 \\
    \hline
    Stagewise training & 52.0 \\
    End-to-End training (\ac{BIER}) & 55.3  \\
    \hline
    \end{tabular}
\end{table}

\section{General Applicability}
\label{sec:general-applicability}

Ideally, our idea of boosting several independent classifiers with a shared
feature representation should be applicable beyond the task of metric learning.
To analyse the generalization capabilities of our method on regular image
classification tasks, we run an experiment on the CIFAR-10~\cite{krizhevsky2009learning} dataset.
CIFAR-10 consists of $60,000$ color images grouped into $10$ categories. Images are of size $32\times32$ pixel.
The dataset is divided into $10,000$ test images and $50,000$ training images. In our experiments we split
the training set into $10,000$ validation images and $40,000$ training images. We select the number of groups for \ac{BIER} based on the performance on the validation set.

The main
objective of this experiment is not to show that we can achieve
state-of-the-art accuracy on CIFAR-10~\cite{krizhevsky2009learning}, but rather to demonstrate that it 
is generally possible to
improve a \ac{CNN} with our method. 
To this end, we run experiments on the CIFAR-10-Quick~\cite{jia2014caffe} and an enlarged version of the CIFAR-10-Quick architecture~\cite{cogswell2016reducing} (see Table~\ref{tbl:network-architecture}). 
In the enlarged version, denoted as CIFAR-10-Quick-Wider, the number of convolution channels 
and the number of neurons in the fully connected layer is doubled. Further, an additional fully connected layer is inserted into the network. 
In both architectures, each convolution layer is
followed by \ac{ReLU} nonlinearity and a pooling layer of size $3\times3$ with
stride $2$. The last fully connected layer in both architectures has no nonlinearity. 

To apply our method,
we divide the last fully
connected layer into $2$ and $4$ non-overlapping groups for the CIFAR-10-Quick and CIFAR-10-Quick-Wider architecture, respectively,
and append a classifier to each group (see Table~\ref{tbl:network-architecture}). As loss function we use crossentropy.
During training time, we apply either our Activation Loss, or Adversarial loss as auxiliary loss function to the last hidden layer of the network.
This encourages the groups to be independent of each other.  
The main reason we have to add the loss function during training time is that
weights change too drastically in networks trained from scratch compared
to fine-tuning a network from a pre-trained ImageNet model. 
Hence, for this type of problems it is more effective to 
additionally encourage diversity of the learners
with a separate loss function.
Further, as opposed to our metric learning version of the algorithm, we can backpropagate the error 
of our auxiliary loss functions to all \ac{CNN} layers without ending up with a trivial solution, where all weights are $\boldsymbol{0}$.

We compare our method to dropout~\cite{srivastava14dropout} applied to the last hidden layer of the network.
As we see in Tables~\ref{tbl:cifar-10-quick-results} and~\ref{tbl:cifar-10-quick-wide-results}, \ac{BIER} with any of our two proposed loss functions improves on the CIFAR-10-Quick architecture over a baseline
with just weight decay by $3.00$\% and over dropout by $1.10$\%. On the larger network which is more prone to overfitting, \ac{BIER} improves over the baseline
by $2.54$\% and over dropout by $1.52$\%.

These preliminary results indicate that \ac{BIER} generalizes well for other
tasks beyond metric learning. Thus, we will further investigate the benefits of
\ac{BIER} for other computer vision tasks in our future work.

\begin{table}[htbp]
    \caption{We use the CIFAR-10-Quick~\cite{jia2014caffe} and an enlarged version of CIFAR-10-Quick~\cite{cogswell2016reducing} architecture.}
    \label{tbl:network-architecture}
    \renewcommand{\arraystretch}{1.3}
    \centering
    \begin{tabular}{|l|l|}
        \hline
        CIFAR-10-Quick           & CIFAR-10-Quick-Wider \\
        \hline
        conv $5\times5\times32$  & conv $5 \times 5 \times 64$ \\
        max-pool $3\times3/2$    & max-pool $3\times3/2$ \\
        \hline
        conv $5\times5\times32$  & conv $5 \times 5 \times 64$ \\
        avg-pool $3\times3/2$    & avg-pool $3\times3/2$ \\
        \hline
        conv $5\times5\times64$  & conv $5 \times 5 \times 128$ \\
        avg-pool $3\times3/2$    & avg-pool $3\times3/2$ \\
        \hline
        fc $64$                  & fc $128$ \\
        \hline
        clf $10\times2$          & fc $128$ \\
        \hline
                                 & clf $10\times4$ \\
        \hline
    \end{tabular}
\end{table}

\begin{table}[htbp]
    \caption{Results on CIFAR-10~\cite{krizhevsky2009learning} with the CIFAR-10-Quick architecture.}
    \label{tbl:cifar-10-quick-results}
    \renewcommand{\arraystretch}{1.3}
    \centering
    \begin{tabular}{ll}
        \hline
        Method   & Accuracy \\
        \hline
        Baseline & 78.72  \\
        Dropout  & 80.62  \\
        \hline
        Activation BIER      & 81.40    \\
        Adversarial BIER     & \textbf{81.72}    \\
        \hline
    \end{tabular}
\end{table}

\begin{table}[htbp]
    \caption{Results on CIFAR-10~\cite{krizhevsky2009learning} with the CIFAR-10-Quick-Wider architecture.}
    \label{tbl:cifar-10-quick-wide-results}
    \centering
    \renewcommand{\arraystretch}{1.3}
    \begin{tabular}{ll}
        \hline
        Method   & Accuracy \\
        \hline
        Baseline & 80.67  \\
        Dropout  & 81.69 \\
        \hline
        Activation BIER     & 83.10    \\
        Adversarial BIER    & \textbf{83.21}    \\
        \hline
    \end{tabular}
\end{table}

\section{Qualitative Comparison of Embeddings}
\label{sec:qualitative-comparison-of-embeddings}

\begin{figure*}[htbp]
    \begin{center}
    \begin{minipage}[t]{0.22\textwidth}
    \centering
    Query
    \end{minipage}
    \hspace{2pt}
    \begin{minipage}[t]{0.22\textwidth}
    \centering
    Learner-1
    \end{minipage}
    \hspace{4pt}
    \begin{minipage}[t]{0.22\textwidth}
    \centering
    Learner-2
    \end{minipage}
    \hspace{4pt}
    \begin{minipage}[t]{0.22\textwidth}
    \centering
    Learner-3
    \end{minipage}
    \hspace{2pt}

    \includegraphics[width=0.22\textwidth]{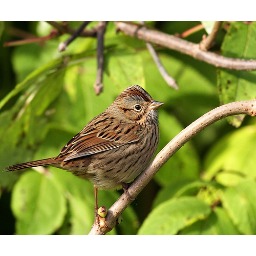}
    \borderred{\includegraphics[width=0.22\textwidth]{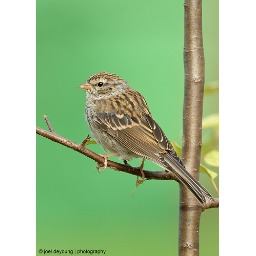}}
    \bordergreen{\includegraphics[width=0.22\textwidth]{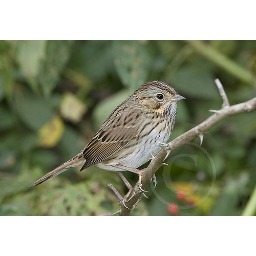}}
    \bordergreen{\includegraphics[width=0.22\textwidth]{figures/qualitative-results/out-cub-learners-adversarial/sample-1423/retrieved--learner-1-label-124-status-ok.jpg}}

    \vspace{0.2cm}
    \includegraphics[width=0.22\textwidth]{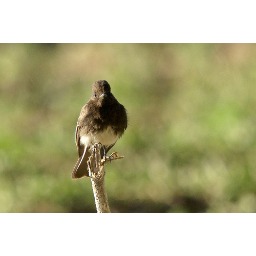}
    \borderred{\includegraphics[width=0.22\textwidth]{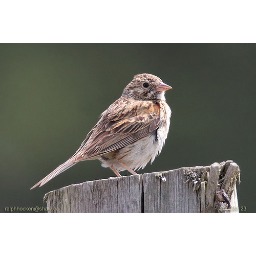}}
    \borderred{\includegraphics[width=0.22\textwidth]{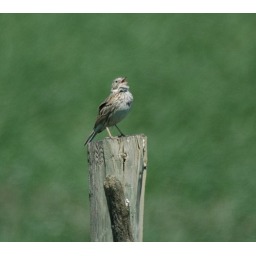}}
    \bordergreen{\includegraphics[width=0.22\textwidth]{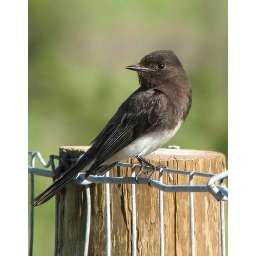}}

    \vspace{0.2cm}
    \includegraphics[width=0.22\textwidth]{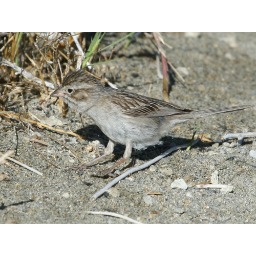}
    \borderred{\includegraphics[width=0.22\textwidth]{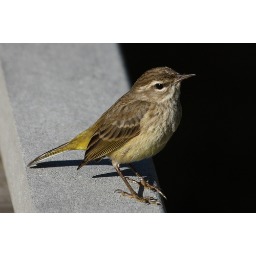}}
    \bordergreen{\includegraphics[width=0.22\textwidth]{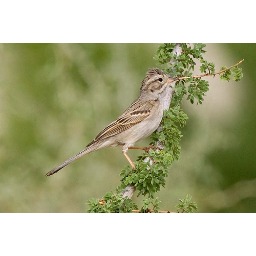}}
    \bordergreen{\includegraphics[width=0.22\textwidth]{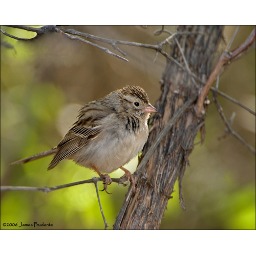}}

    \end{center}

    \caption{Qualitative results on the CUB-200-2011~\cite{WahCUB_200_2011} dataset of the different learners in our ensemble. 
             We retrieve the most similar image to the query image for learner 1, 2 and 3, respectively.
             Correct results are highlighted \textcolor{green}{green} and incorrect results are highlighted \textcolor{red}{red}.}
    \label{fig:qualitative-cub-learners}
\end{figure*}

To illustrate the differences between the learned embeddings we show several qualitative examples in Figure~\ref{fig:qualitative-cub-learners}. 
Successive learners typically perform better at harder examples compared to previous learners, which have a smaller embedding size.

\section{Qualitative Results}
\label{sec:qualitative-results}

\begin{figure*}[htbp]
    \begin{center}
    \begin{minipage}{0.16\textwidth}
    \centering
    Query
    \end{minipage}
    \begin{minipage}{0.16\textwidth}
    \centering
    1
    \end{minipage}
    \begin{minipage}{0.16\textwidth}
    \centering
    2
    \end{minipage}
    \begin{minipage}{0.16\textwidth}
    \centering
    3
    \end{minipage}
    \begin{minipage}{0.16\textwidth}
    \centering
    4
    \end{minipage}
    \begin{minipage}{0.16\textwidth}
    \centering
    5
    \end{minipage}
    \hspace{0.5cm}

    \includegraphics[width=0.16\textwidth]{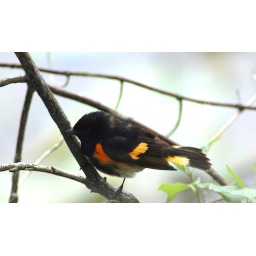}
    \includegraphics[width=0.16\textwidth]{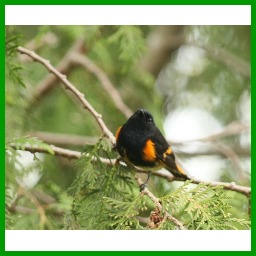}
    \includegraphics[width=0.16\textwidth]{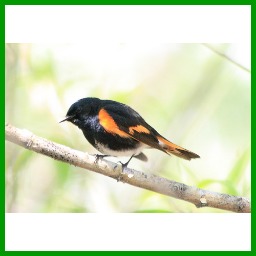}
    \includegraphics[width=0.16\textwidth]{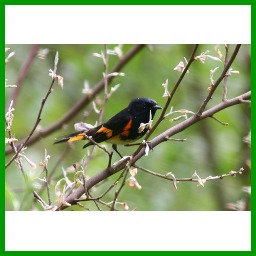}
    \includegraphics[width=0.16\textwidth]{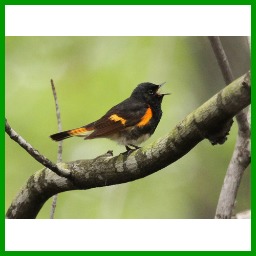}
    \includegraphics[width=0.16\textwidth]{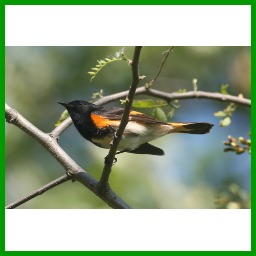}

    \vspace{0.2cm}
    \includegraphics[width=0.16\textwidth]{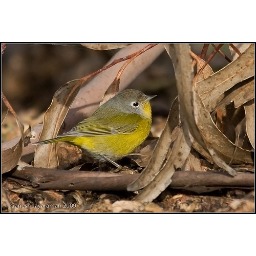}
    \includegraphics[width=0.16\textwidth]{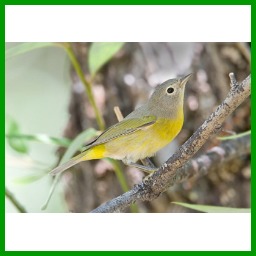}
    \includegraphics[width=0.16\textwidth]{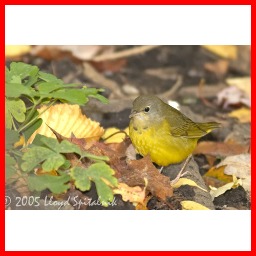}
    \includegraphics[width=0.16\textwidth]{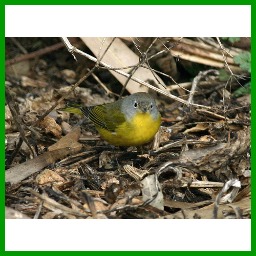}
    \includegraphics[width=0.16\textwidth]{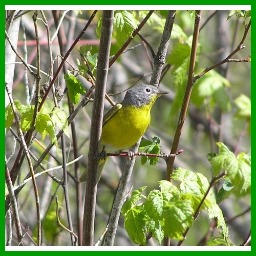}
    \includegraphics[width=0.16\textwidth]{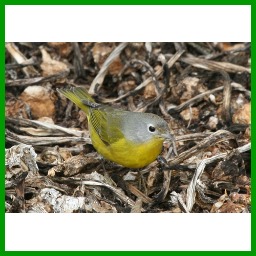}

    \vspace{0.2cm}
    \includegraphics[width=0.16\textwidth]{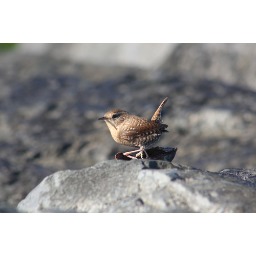}
    \includegraphics[width=0.16\textwidth]{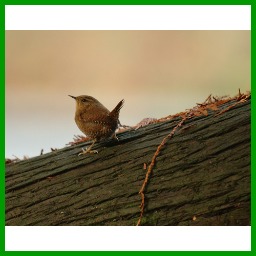}
    \includegraphics[width=0.16\textwidth]{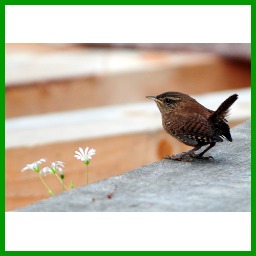}
    \includegraphics[width=0.16\textwidth]{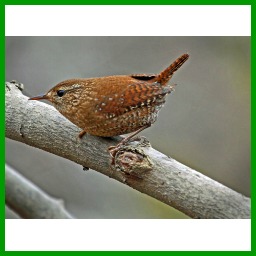}
    \includegraphics[width=0.16\textwidth]{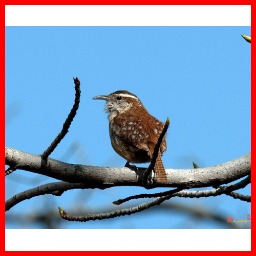}
    \includegraphics[width=0.16\textwidth]{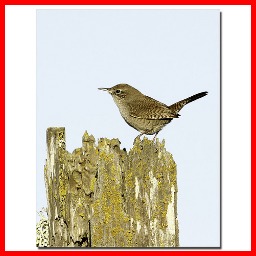}

    \end{center}
    \caption{Qualitative results on the CUB-200-2011~\cite{WahCUB_200_2011} dataset. 
             We retrieve the 5 most similar images to the query image.
             Correct results are highlighted \textcolor{green}{green} and incorrect results are highlighted \textcolor{red}{red}.}
    \label{fig:qualitative-cub}
\end{figure*}

\begin{figure*}[htbp]
    \begin{center}
    \begin{minipage}{0.16\textwidth}
    \centering
    Query
    \end{minipage}
    \begin{minipage}{0.16\textwidth}
    \centering
    1
    \end{minipage}
    \begin{minipage}{0.16\textwidth}
    \centering
    2
    \end{minipage}
    \begin{minipage}{0.16\textwidth}
    \centering
    3
    \end{minipage}
    \begin{minipage}{0.16\textwidth}
    \centering
    4
    \end{minipage}
    \begin{minipage}{0.16\textwidth}
    \centering
    5
    \end{minipage}
    \hspace{0.5cm}

    \includegraphics[width=0.16\textwidth]{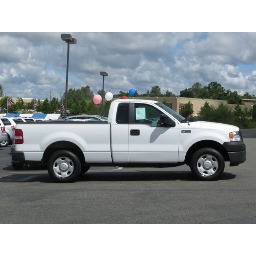}
    \includegraphics[width=0.16\textwidth]{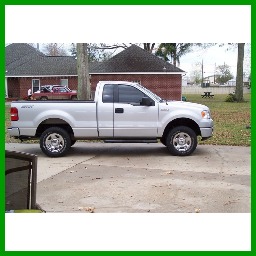}
    \includegraphics[width=0.16\textwidth]{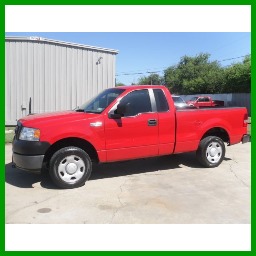}
    \includegraphics[width=0.16\textwidth]{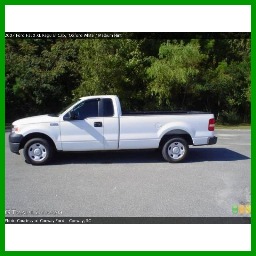}
    \includegraphics[width=0.16\textwidth]{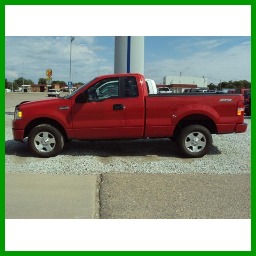}
    \includegraphics[width=0.16\textwidth]{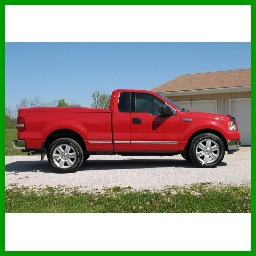}

    \vspace{0.2cm}
    \includegraphics[width=0.16\textwidth]{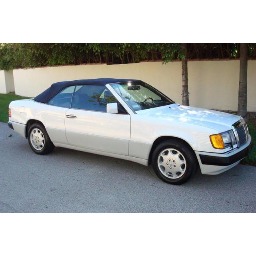}
    \includegraphics[width=0.16\textwidth]{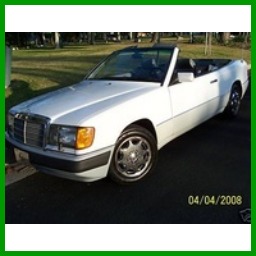}
    \includegraphics[width=0.16\textwidth]{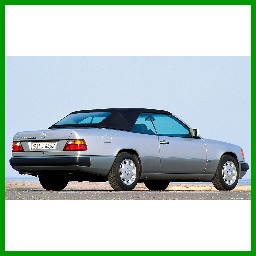}
    \includegraphics[width=0.16\textwidth]{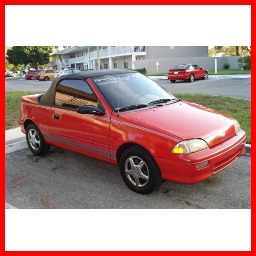}
    \includegraphics[width=0.16\textwidth]{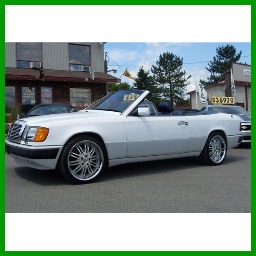}
    \includegraphics[width=0.16\textwidth]{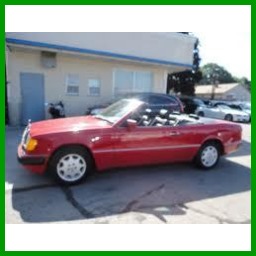}

    \vspace{0.2cm}
    \includegraphics[width=0.16\textwidth]{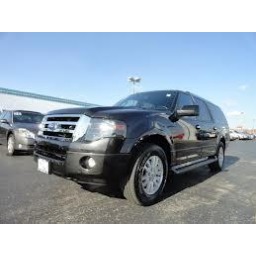}
    \includegraphics[width=0.16\textwidth]{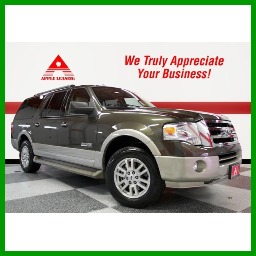}
    \includegraphics[width=0.16\textwidth]{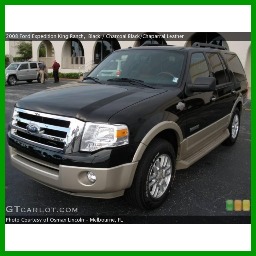}
    \includegraphics[width=0.16\textwidth]{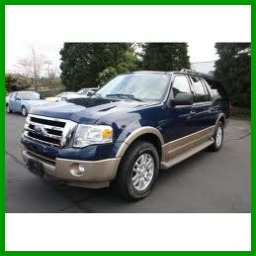}
    \includegraphics[width=0.16\textwidth]{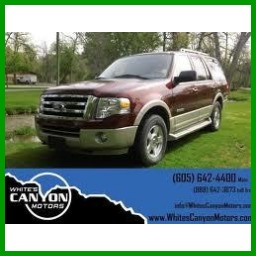}
    \includegraphics[width=0.16\textwidth]{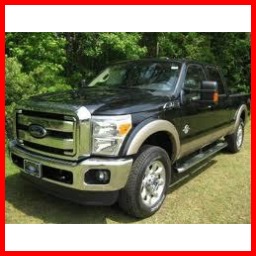}

    \end{center}
    \caption{Qualitative results on the Cars-196~\cite{krause20133d} dataset. 
             We retrieve the 5 most similar images to the query image.
             Correct results are highlighted \textcolor{green}{green} and incorrect results are highlighted \textcolor{red}{red}.}
    \label{fig:qualitative-cars}
\end{figure*}

\begin{figure*}[htbp]
    \begin{center}
    \begin{minipage}{0.16\textwidth}
    \centering
    Query
    \end{minipage}
    \begin{minipage}{0.16\textwidth}
    \centering
    1
    \end{minipage}
    \begin{minipage}{0.16\textwidth}
    \centering
    2
    \end{minipage}
    \begin{minipage}{0.16\textwidth}
    \centering
    3
    \end{minipage}
    \begin{minipage}{0.16\textwidth}
    \centering
    4
    \end{minipage}
    \begin{minipage}{0.16\textwidth}
    \centering
    5
    \end{minipage}
    \hspace{0.5cm}

    \includegraphics[width=0.16\textwidth]{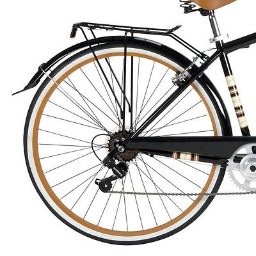}
    \includegraphics[width=0.16\textwidth]{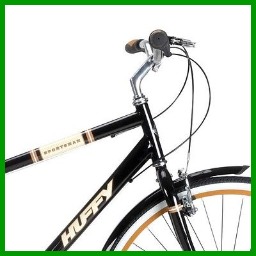}
    \includegraphics[width=0.16\textwidth]{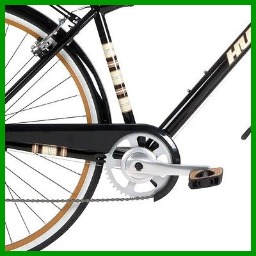}
    \includegraphics[width=0.16\textwidth]{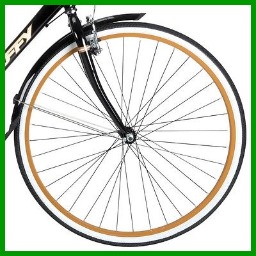}
    \includegraphics[width=0.16\textwidth]{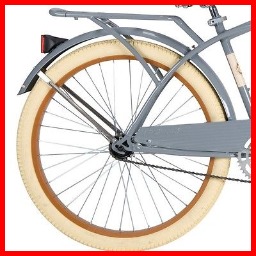}
    \includegraphics[width=0.16\textwidth]{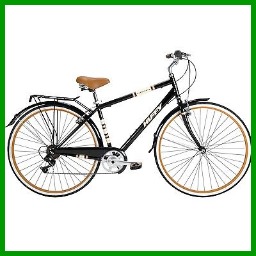}

    \vspace{0.2cm}
    \includegraphics[width=0.16\textwidth]{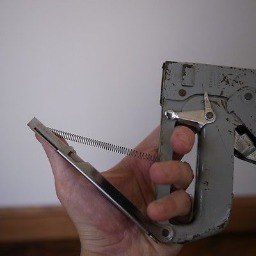}
    \includegraphics[width=0.16\textwidth]{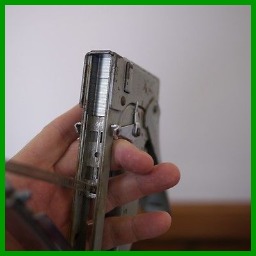}
    \includegraphics[width=0.16\textwidth]{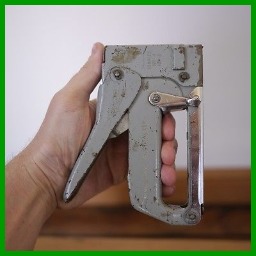}
    \includegraphics[width=0.16\textwidth]{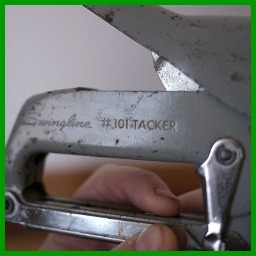}
    \includegraphics[width=0.16\textwidth]{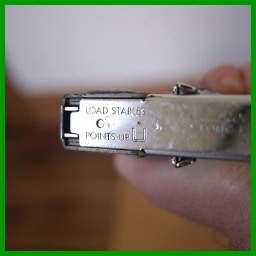}
    \includegraphics[width=0.16\textwidth]{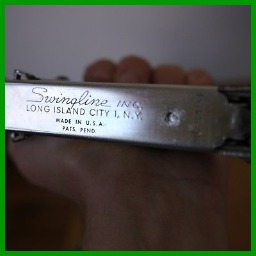}

    \vspace{0.2cm}
    \includegraphics[width=0.16\textwidth]{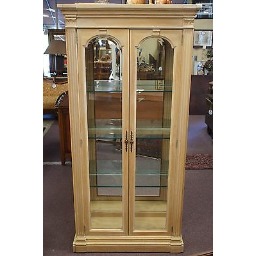}
    \includegraphics[width=0.16\textwidth]{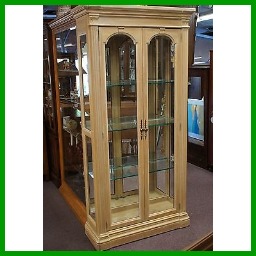}
    \includegraphics[width=0.16\textwidth]{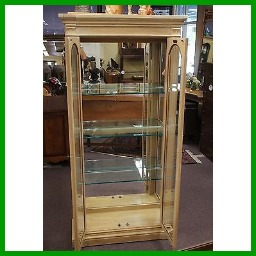}
    \includegraphics[width=0.16\textwidth]{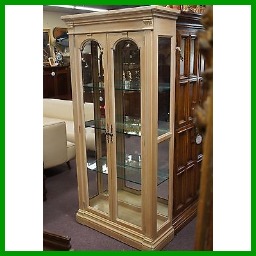}
    \includegraphics[width=0.16\textwidth]{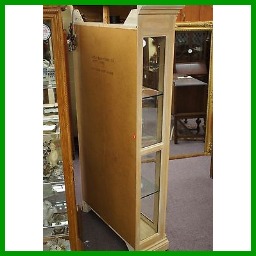}
    \includegraphics[width=0.16\textwidth]{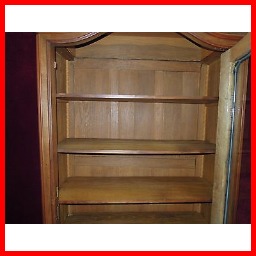}

    \end{center}
    \caption{Qualitative results on the Stanford Online Products~\cite{oh2016deep} dataset. 
             We retrieve the 5 most similar images to the query image.
             Correct results are highlighted \textcolor{green}{green} and incorrect results are highlighted \textcolor{red}{red}.}
    \label{fig:qualitative-products}
\end{figure*}

\begin{figure*}[htbp]
    \begin{center}
    \begin{minipage}{0.16\textwidth}
    \centering
    Query
    \end{minipage}
    \begin{minipage}{0.16\textwidth}
    \centering
    1
    \end{minipage}
    \begin{minipage}{0.16\textwidth}
    \centering
    2
    \end{minipage}
    \begin{minipage}{0.16\textwidth}
    \centering
    3
    \end{minipage}
    \begin{minipage}{0.16\textwidth}
    \centering
    4
    \end{minipage}
    \begin{minipage}{0.16\textwidth}
    \centering
    5
    \end{minipage}
    \hspace{0.5cm}

    \includegraphics[width=0.16\textwidth]{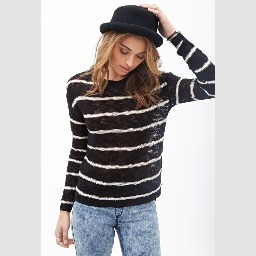}
    \includegraphics[width=0.16\textwidth]{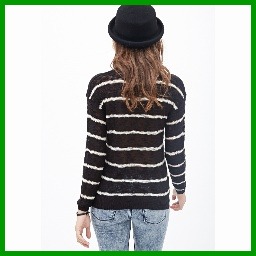}
    \includegraphics[width=0.16\textwidth]{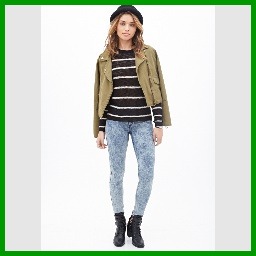}
    \includegraphics[width=0.16\textwidth]{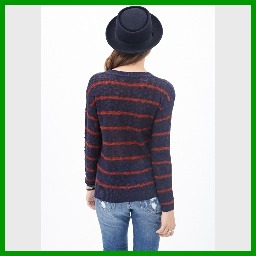}
    \includegraphics[width=0.16\textwidth]{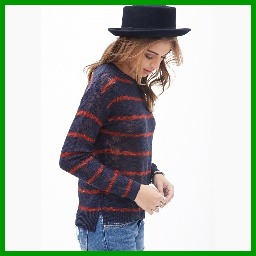}
    \includegraphics[width=0.16\textwidth]{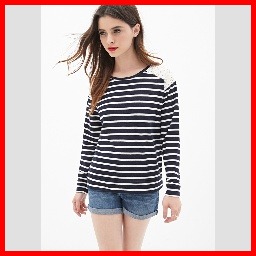}

    \vspace{0.2cm}
    \includegraphics[width=0.16\textwidth]{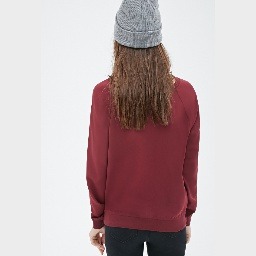}
    \includegraphics[width=0.16\textwidth]{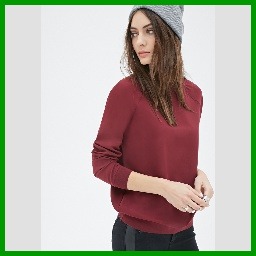}
    \includegraphics[width=0.16\textwidth]{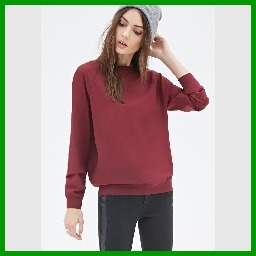}
    \includegraphics[width=0.16\textwidth]{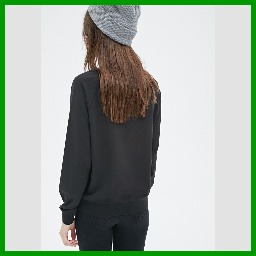}
    \includegraphics[width=0.16\textwidth]{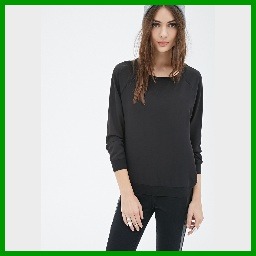}
    \includegraphics[width=0.16\textwidth]{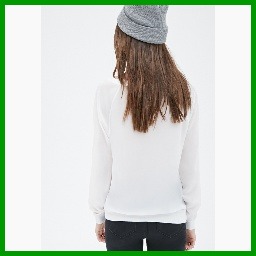}

    \vspace{0.2cm}
    \includegraphics[width=0.16\textwidth]{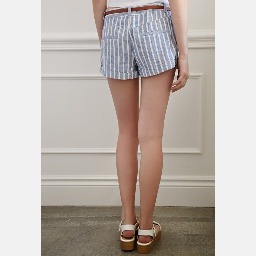}
    \includegraphics[width=0.16\textwidth]{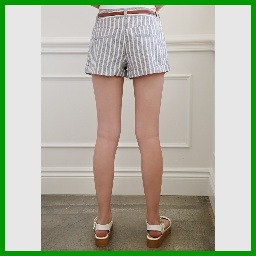}
    \includegraphics[width=0.16\textwidth]{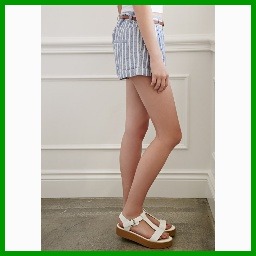}
    \includegraphics[width=0.16\textwidth]{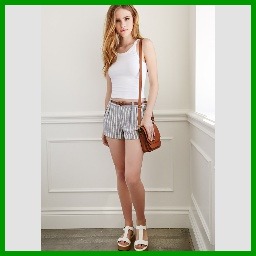}
    \includegraphics[width=0.16\textwidth]{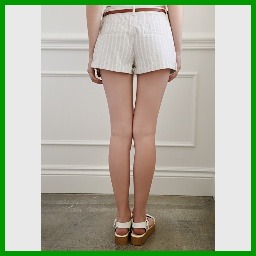}
    \includegraphics[width=0.16\textwidth]{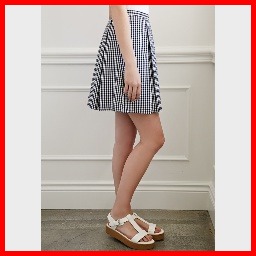}

    \end{center}
    \caption{Qualitative results on the In-Shop Clothes Retrieval~\cite{liu2016deepfashion} dataset. 
             We retrieve the 5 most similar images to the query image.
             Correct results are highlighted \textcolor{green}{green} and incorrect results are highlighted \textcolor{red}{red}.}
    \label{fig:qualitative-clothes}
\end{figure*}

\begin{figure*}[htbp]
    \begin{center}
    \begin{minipage}{0.16\textwidth}
    \centering
    Query
    \end{minipage}
    \begin{minipage}{0.16\textwidth}
    \centering
    1
    \end{minipage}
    \begin{minipage}{0.16\textwidth}
    \centering
    2
    \end{minipage}
    \begin{minipage}{0.16\textwidth}
    \centering
    3
    \end{minipage}
    \begin{minipage}{0.16\textwidth}
    \centering
    4
    \end{minipage}
    \begin{minipage}{0.16\textwidth}
    \centering
    5
    \end{minipage}
    \hspace{0.5cm}

    \includegraphics[width=0.16\textwidth]{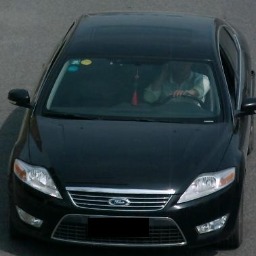}
    \includegraphics[width=0.16\textwidth]{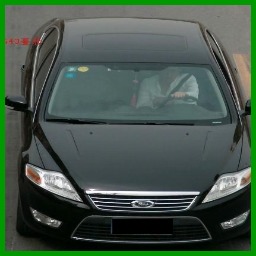}
    \includegraphics[width=0.16\textwidth]{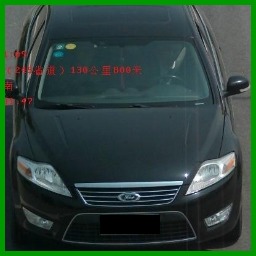}
    \includegraphics[width=0.16\textwidth]{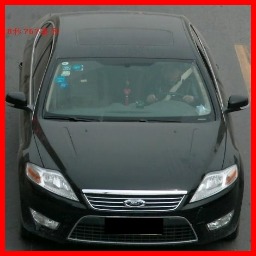}
    \includegraphics[width=0.16\textwidth]{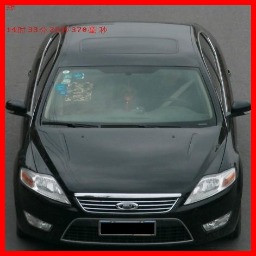}
    \includegraphics[width=0.16\textwidth]{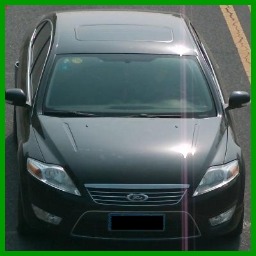}

    \vspace{0.2cm}
    \includegraphics[width=0.16\textwidth]{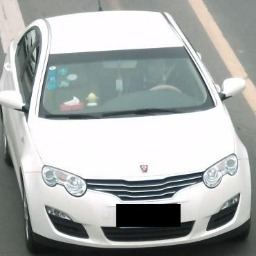}
    \includegraphics[width=0.16\textwidth]{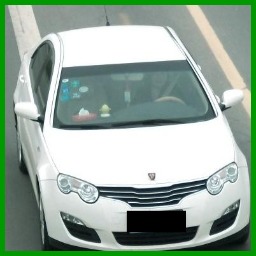}
    \includegraphics[width=0.16\textwidth]{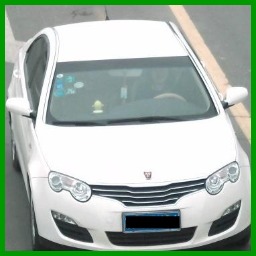}
    \includegraphics[width=0.16\textwidth]{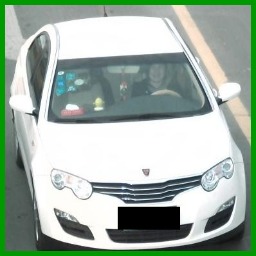}
    \includegraphics[width=0.16\textwidth]{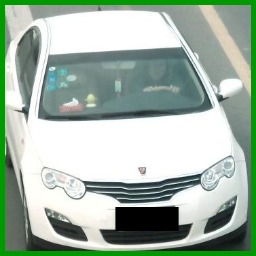}
    \includegraphics[width=0.16\textwidth]{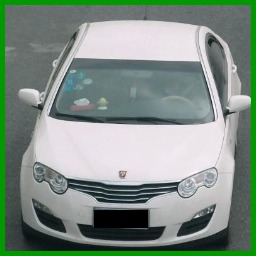}

    \vspace{0.2cm}
    \includegraphics[width=0.16\textwidth]{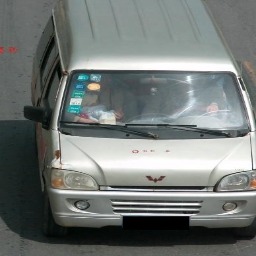}
    \includegraphics[width=0.16\textwidth]{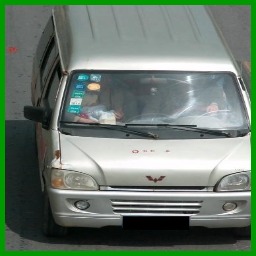}
    \includegraphics[width=0.16\textwidth]{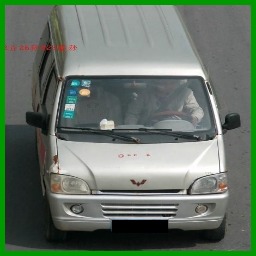}
    \includegraphics[width=0.16\textwidth]{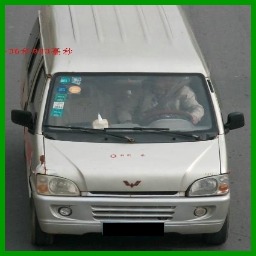}
    \includegraphics[width=0.16\textwidth]{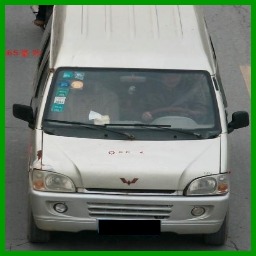}
    \includegraphics[width=0.16\textwidth]{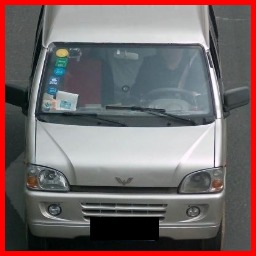}

    \end{center}
    \caption{Qualitative results on the VehicleID~\cite{liu2016deep} dataset. 
             We retrieve the 5 most similar images to the query image.
             Correct results are highlighted \textcolor{green}{green} and incorrect results are highlighted \textcolor{red}{red}.}
    \label{fig:qualitative-vehicleid}
\end{figure*}

To illustrate the effectiveness of \ac{BIER} we show some qualitative examples in Figures~\ref{fig:qualitative-cub},~\ref{fig:qualitative-cars},~\ref{fig:qualitative-products},~\ref{fig:qualitative-clothes} and~\ref{fig:qualitative-vehicleid}.

\ifCLASSOPTIONcompsoc
  \section*{Acknowledgments} 

\else
  \section*{Acknowledgment}
\fi

This project was supported by the Austrian Research Promotion Agency (FFG) projects MANGO (836488) and Darknet (858591). We gratefully acknowledge the support of NVIDIA Corporation with the donation of GPUs used for this research.

\ifCLASSOPTIONcaptionsoff
  \newpage
\fi

\bibliographystyle{IEEEtran}
\bibliography{abbreviation_short,egbib}

\end{document}